\documentclass[multiauthors,onecolumn]{LaTeXclass/cohere}

\usepackage{amsmath,amsfonts,bm}

\def\eqref#1{equation~\ref{#1}}

\def\1{\bm{1}}

\DeclareMathAlphabet{\mathsfit}{\encodingdefault}{\sfdefault}{m}{sl}
\SetMathAlphabet{\mathsfit}{bold}{\encodingdefault}{\sfdefault}{bx}{n}

\DeclareMathOperator*{\argmax}{arg\,max}

\usepackage{floatrow}

\usepackage{hyperref}
\usepackage{url}
\usepackage{comment}
\usepackage{capt-of}
\usepackage{cleveref}
\usepackage{tcolorbox}
\usepackage{subcaption}
\usepackage{graphicx}
\usepackage{wrapfig}   
\usepackage{sidecap}   
\usepackage{booktabs}  
\usepackage{multirow}    
\usepackage{array}     
\usepackage{float}     
\usepackage{caption}   
\usepackage[table]{xcolor} 
\usepackage{amssymb}
\usepackage{makecell} 
\usepackage{multicol}
\usepackage{amsmath}

\newfloatcommand{capbtabbox}{table}[][\FBwidth]

\newcommand{\bon}{\textsc{BoN}}
\newcommand{\fusion}{\textsc{FusioN}}
\newcommand{\aya}{\textsc{Aya Expanse 8B}}
\newcommand{\command}{\textsc{Command A}}
\newcommand{\kimi}{\textsc{Kimi-K2-Instruct}}

\newcommand{\gemmabig}{\textsc{Gemma3-27B-It}}

\newcommand{\gemmasmall}{\textsc{Gemma3-4B-It}}

\newcommand{\deepseek}{\textsc{DeepSeek-V3}}
\newcommand{\qwenbig}{\textsc{Qwen3-235B}}
\newcommand{\ourrm}{\textsc{RM}}

\newcommand{\geminipro}{\textsc{Gemini2.5-Pro}}
\newcommand{\geminiflash}{\textsc{Gemini2.5-Flash}}
\newcommand{\gpt}{\textsc{GPT-4o}}
\newcommand{\comet}{\textsc{XCometXL}}

\setcitestyle{number,square}

\title{Making, not Taking, the Best of N}

\author{name={Ammar Khairi\fa},affiliation={1}}
\author{name={Daniel D'souza},affiliation={1}}
\author{name={Marzieh Fadaee},affiliation={1}}
\author{name={Julia Kreutzer\psa},affiliation={1}}
\affiliations{\item[1] Cohere Labs}

\corresponding[*]{\{\url{ammar}, \url{juliakreutzer}\}\url{{@cohere.com}}}

\abstract{
\justifying
Obtaining high-quality generations in modern LLMs has largely been framed as a selection problem: identifying a single \textit{winning} generation from a diverse pool of $N$ samples, the Best-of-$N$ (\bon{}).
Yet, this approach is inherently zero-sum, discarding diverse and potentially useful information from the pool. Instead, we explore a collaborative setup, where all candidates can potentially contribute to the final \textit{winning} generation. To this end, we propose \textbf{Fusion-of-$N$} (\fusion{}): a method that uses a general LLM judge to synthesize the most informative elements of each sample into a single final answer. 
We compare \fusion{} to \bon{} in two settings, (i) \textbf{test-time scaling}, where we sample and aggregate from a single model at test-time (ii) \textbf{synthetic data generation}, where we fuse samples from a pool of diverse teachers to improve a student model. We extensively benchmark both setups across 11 languages, 3 diverse tasks and varying model scales. Across the bench, \fusion{} consistently outperforms \bon{} showing versatility and robustness both in test-time scaling and in downstream gains from synthetic data generation. We also perform extensive analysis on \fusion{}, where it shows surprising strengths and robustness under challenging settings.
These results show that we should shift how we think about evaluating and utilizing LLM generations from a monolithic measure of quality, to embracing their polylithic nature. This shift allows us to integrate diverse strengths, unlock latent potential, and achieve improvements that were previously inaccessible through selection alone.
}

\renewcommand{\FONTauthors}{\bfseries\Large}
\renewcommand{\FONTauthor}{\bfseries\Large}
\begin{document}

\section{Introduction}

Many of today's advances in LLMs rely heavily on aggregation at inference: The dominant approach, Best-of-$N$ (\bon{}), involves generating multiple candidates and selecting one among them as the final output.  
This approach has proven highly effective for test-time scaling in tasks ranging from math reasoning and translation to open-ended tasks ~\citep{Snell2025ScalingLT, khairi2025lifegivessamplesbenefits, yao2023tree, wang2023selfconsistency}, and for producing synthetic data used in fine-tuning~\citep{jayalath2025computeteacherturninginference,muennighoff2025s1simpletesttimescaling}, especially in multilingual setups~\citep{grattafiori2024llama3herdmodels,dang2024ayaexpansecombiningresearch, martins2025eurollm9btechnicalreport, hernándezcano2025apertusdemocratizingopencompliant,lai-nissim-2024-mcot,hwang2025learngloballyspeaklocally,odumakinde-etal-2025-multilingual,rei2025towerbridginggeneralitytranslation}. 
However, existing aggregation methods treat generations as competitors in a \textit{zero-sum game}. 
Whether through majority voting~\citep{brown2024largelanguagemonkeysscaling}, self-consistency ~\citep{wang2023selfconsistency}, or reward-model scoring~\citep{NEURIPS2022_b1efde53}, the goal is to find the single best answer while discarding the rest.
This hard \textbf{selection} step imposes clear limitations: it discards the diversity of reasoning paths that could be combined to produce stronger answers. 
It wastes much of the compute spent generating samples and risks reward hacking~\citep{skalse2022defining,NEURIPS2022_3d719fee,ichihara2025evaluation}: the candidate that maximizes a judge's score is not always the most correct or useful.

In today's fast-shifting LLM landscape, where leaderboard wins change hands quickly, treating quality as a single monolithic dimension is increasingly outdated.
In practice, there is rarely a single ``best'' answer; diverse outputs often complement one another.
This motivates our central question: \textit{can we go beyond selection and design a method that makes fuller use of all generated samples?} 
We propose \fusion{}, a simple synthesis-based alternative to \bon{} that exploits the generative abilities of LLMs to integrate complementary signals across candidates---truly making, rather than merely taking, the best of $N$.

\begin{figure}[t]
   \centering
   \includegraphics[width=\textwidth]{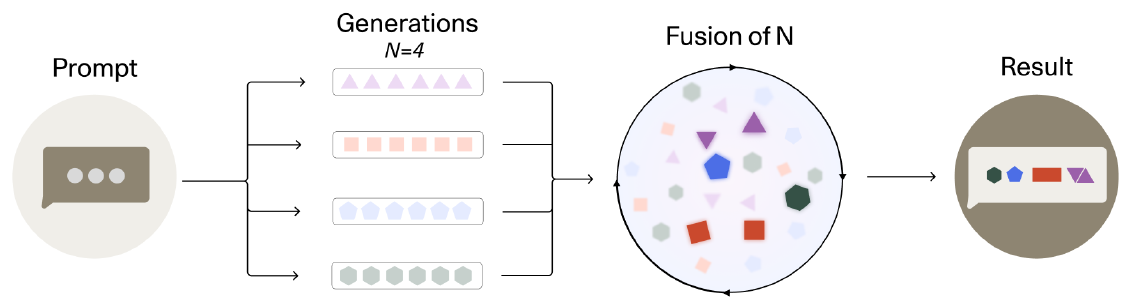}
   \caption{\fusion{} principle: Multiple generations (here $N=4$, from one or multiple models) get fused into one final generation combining the strengths of each individual generation.
   }
   \label{fig:fusion_overview}
\end{figure}

We treat aggregation as a \textbf{synthesis problem} rather than a selection problem. \Cref{fig:fusion_overview} illustrates our idea: We use a strong LLM judge, the fusor, to integrate the complementary strengths of multiple candidates into a single answer. Our proposed Fusion-of-$N$ (\fusion{}) method is simple, general, versatile and can directly replace \bon{} with no modifications beyond access to a reasonably strong generative LLM that acts as a fusor. The \textit{polylithic} understanding of quality allows us to  decompose complex problems into  compositional ones that are more tractable. \fusion{} optimizes across samples and integrates complementary insights into a single, higher-quality answer. Going beyond the initial sample pool is especially valuable when the pool is strong and diverse, and for problems that naturally benefit from diversity.
Intuitively, this mirrors how experts synthesize knowledge from multiple domains and perspectives.

We perform a comprehensive evaluation of \fusion{} as a replacement for \bon{} across test-time scaling and data generation: For \textbf{test-time scaling}, we measure the effectiveness of \fusion{} with multiple samples from 8B and 111B models on open-ended generation and machine translation tasks. We evaluate the \textbf{impact of synthetic data} generated with \fusion{} in terms of data quality and downstream results after fine-tuning a 7B and a 111B model on open-ended prompts, math and factual reasoning tasks. In both setups our evaluations are spanning multiple languages to test \fusion{} under diverse and challenging conditions.

Our results show that synthesis is not only more effective, but also more sample-efficient: 
\fusion{} consistently outperforms \bon{} under the same sampling budget, and in some cases even surpasses the oracle, revealing that selection is not the upper bound. 
It proves robust under weaker teacher pools, showing that diversity can be leveraged even when individual contributors are limited. 
We observe that fine-tuning on \fusion{} data enables models to outperform even the strongest single teacher, showing that synthesis distills collective knowledge in ways that selection cannot.
Finally, our analysis provides the first detailed look into the mechanisms of synthesis, uncovering both its strengths---sample efficiency, robustness, and adaptability---and its limits on tightly constrained math tasks. To summarize, our contributions are:

\textbf{Conceptual shift from selection to synthesis.} We present compelling evidence for reframing aggregation as synthesis problem. In contrast to previous works in this direction (\cref{sec:related}), \fusion{} is simple, easily customizable and works out-of-the-box, making it an attractive substitute for \bon{}. 

\textbf{Demonstrated gains across test-time scaling and data generation.} \fusion{} consistently outperforms \bon{} in both settings where candidate aggregation is used today: (i) \textit{test-time scaling}, where it yields substantial improvements (e.g., +3.8\% win-rate vs \geminipro{} on mArena-v2, +3.7 \comet{} on translation), and (ii) \textit{synthetic data generation}, where it produces higher-quality datasets that drive downstream gains across diverse tasks (+2.5\% on mArena-v2 vs \geminiflash{}, +0.8 on WMT, +1.8\% on GeoFactX answer accuracy and +1.1\% on reasoning quality).

\textbf{Robustness and efficiency across models and settings.} Our analysis shows that \fusion{} is more sample-efficient and robust than \bon{}. It maintains high performance with smaller or weaker teacher pools, benefits from larger fusor models, and scales effectively with added test-time compute. These properties make it a practical and generalizable approach for both test-time scaling and synthetic data generation, even under constrained or imperfect conditions.

This work redefines how we measure and leverage LLM outputs. 
Instead of treating generations as isolated candidates, we embrace their diversity and complementary strengths, synthesizing them into more powerful, coherent results. 
Our findings show that treating LLMs as collaborators and not competitors unlocks higher-quality outputs and more impact on downstream use cases, pointing toward a fundamentally more effective paradigm for large-scale language model deployment.

\section{Methodology: From Selection to Synthesis}\label{sec:methods}

\textbf{Selection with Best-of-N} (\bon{}) Given a prompt \( x \), a pool of candidates \( y\in Y \), and a scoring function \( S \), the \bon{} method selects the optimal candidate \( y^* \) by maximizing a scalar score:
\[ y^* = \argmax_{y \in Y} S(y, x) \]
The scoring function could be a specialized reward model as used in rejection sampling for synthetic data generation~\citep{grattafiori2024llama3herdmodels}, or test-time scaling~\citep{cobbe2021training}.
The score could also be produced by a generative LLM that is prompted to predict a scalar score~\citep{kim-etal-2024-prometheus}, though in practice trained classifiers often perform better, for instance many top models on RewardBench~\citep{malik2025rewardbench2advancingreward} leaderboard are sequence classifiers.
These type of scoring functions are typically optimized on verifiable domains and pairwise human preferences~\citep{cobbe2021training,NEURIPS2022_b1efde53}.

\textbf{Limitations of \bon{}} The limiting factors for selection with \bon{} are (1) the alignment with the desired task~\citep{lambert2020objective,pan2022the,ichihara2025evaluation,viswanathan2025checklistsbetterrewardmodels}, (2) and the quality of the generated sample pool (as per definition, the final generation can \textit{only be as good as the best} of the candidates). 
For domains with verifiable problems, the alignment can easily be improved by scaling up training data for the reward model~\citep{liu-etal-2025-acemath}, but even with expensive ensembles~\citep{eisenstein2024helping} risks of overfitting to an imperfect proxy remain~\citep{stroebl2024inferencescalingflawslimits}. Scaling up reward model alignment is less transferrable to open domains like chat or open-ended question answering, where this signal needs to be obtained from human feedback~\citep{huang2025is,viswanathan2025checklistsbetterrewardmodels}.
Similarly, a poor initial sample pool can be improved by diversification~\citep{chen2025trulyneedsamplesmultillm} or optimized sampling~\citep{khairi2025lifegivessamplesbenefits}, or simply scaling up the number of samples, sometimes requiring thousands of samples for test-time scaling to be effective~\citep{stroebl2024inferencescalingflawslimits,brown2024largelanguagemonkeysscaling}, 
which makes it extremely resource-intensive.

\textbf{Synthesis with Fusion-of-N} (\fusion{}) A fusor model \( F \) (a standard LLM) generates a \textit{new} response \( y^{\star} \) based on the input prompt \( X \), and a pool of candidates \( Y \): 
\[ y^{\star} = F(x, Y),\quad y^*\notin Y  \] 
This means that the final generation $y^*$ is conditionally dependent on the other candidates, and, can---in contrast to \bon{}---\textit{exceed} the original pool in quality (see \cref{sec:analysis}). It can be seen as a form of collaborative refinement: Rather than only selecting a sample according to a monolithic notion of quality, \fusion{} goes beyond and productively \textit{integrates a polylithic notion of quality into the synthesis} of a better sample. The polylithic view, meaning that we acknowledge the existence of higher and lower-quality parts in each sample, is particularly well suited for long generations for complex prompts. \fusion{} can ``mix and match'' fragments of variable size (e.g. tokens, terms, sentences, ...) that stand out in quality in each of the provided samples (see the example in ~\cref{fig:example_fused_generation}). 
\bon{} is captured as a special case: the fusor still has the option to copy one whole generation if it outperforms all others for the entire sequence.

\textbf{Components of \fusion{}} The success of \fusion{} depends on the capabilities of the judge to comparatively evaluate, extract and aggregate the best parts of each generation. We will show in \cref{sec:analysis} that there appears to be a threshold in model size that needs to be crossed for \fusion{} to work without any specialized training. Our analysis also shows that the choice of fusor, given a certain model size, seems less important than the composition of the sample pool.
One major advantage over using a reward model, is that the \fusion{} prompt (ours in \cref{tab:fusion_prompt}) allows for in-context learning and adaptation \emph{without any training}.
It can be tuned to steer \fusion{} behavior in ambiguous cases, such as concerning safety standards (e.g. with a constitution~\citep{bai2022constitutionalaiharmlessnessai}), tone or model identity, and how much it should attempt to integrate parts from all samples or also discard the worst ones entirely. With chain-of-thought prompting~\citep{wei2022chain} or reasoning models as fusors, we also have the possibility to scale up \fusion{} compute where desired. In preliminary experiments we found it important to instruct the model to not only focus on the best, but also discard the worst parts. We have not conducted any prompt tuning beyond that, but practitioners are invited to tune their \fusion{} prompt to their use cases.

\section{Experimental Setup}\label{sec:experiments}
Our experiments span two prominent environments for \bon{}, the first focused on \textbf{test-time scaling}, and the second focused on \textbf{synthetic data generation}.
In both cases, our intervention of replacing \bon{} by \fusion{} is minimal: Both methods receive the \emph{identical set of generations} for the same prompts, but aggregate it differently to produce the final generation.

\subsection{Models for Test-Time Scaling}\label{sec:tts_setup}
We study the test-time scaling behavior for multilingual models of two sizes: \aya{} and \command{} at 111B. We use temperature sampling at $T=0.7$ to generate $N=5$ samples from each model (see \cref{fig:fusion_arena_scaling_tts} for various $N$). 
We use a competitive in-house multilingual Reward Model (\ourrm{})\footnote{It scores an average score of 76.1 on the English RewardBench 2~\citep{malik2025rewardbench2advancingreward}, which at the time of submission (24 Sept 2025), places it at 11th place. On multilingual RewardBench~\citep{gureja-etal-2025-rewardbench} it scores an average of 87.6 across languages, placing it on top of the leaderboard of openly benchmarked models.} for scoring the candidates in \bon{} and \command{} as fusor in \fusion{} (ablation and comparison to \textsc{Gemma} models~\citep{team2025gemma} in \cref{fig:fusor_ablation}).

\subsection{Models and Data for Synthetic Data Generation}\label{sec:sft}
\textbf{Models.} For synthetic data generation, we employ five open and strong models of varying size and families as teachers: \gemmabig{}, \kimi{}, \qwenbig{}, \deepseek{} and \command{} ~\citep{team2025gemma, kimiteam2025kimik2openagentic, yang2025qwen3technicalreport, deepseekai2025deepseekv3technicalreport, cohere2025commandaenterprisereadylarge}.
We sample a low temperature completion ($\tau=0.3$) from each of them to generate the pool of samples for each prompt. 
From this pool, we then select one completion for supervised fine-tuning (SFT), either with \ourrm{} or \command{} as fusor. Ablations regarding pool composition and fusor model choice will follow in ~\cref{tab:ufb_ablations}.
For fine-tuning, we choose an 111B instruction-tuned LLM as our baseline model for our main SFT experiments, and perform an ablation with a smaller 7B Base LLM baseline (\cref{sec:extended_results}). 
Finetuning hyperparameters are listed in \cref{app:hyperparams}. We do not apply test-time scaling on top of our fine-tuned models.

\textbf{General Fine-tuning Dataset.} 
For our main fine-tuning experiments, we randomly sample 10k prompts from UltraFeedback Binarized (UFB) \citep{tunstall2023zephyr}, an English preference dataset with 61k pairs that was previously used to measure the impacts of data composition in fine-tuning~\citep{odumakinde-etal-2025-multilingual,li2025draftsanswersunlockingllm}. 
We translate the prompts automatically
into 9 languages: German, French, Spanish, Chinese, Japanese, Arabic, Korean, Italian, Portuguese.

\textbf{Reasoning Fine-tuning Dataset.} Learning to reason is often approached through synthetic data, where models imitate reasoning traces from a single teacher~\citep{shridhar-etal-2023-distilling,muennighoff2025s1simpletesttimescaling,hwang2025learngloballyspeaklocally}. Here, we apply our \fusion{} approach to learn to reason from multiple teachers.
We add a second, smaller, batch of prompts for domain-specific reasoning tasks: We add the prompts from the GeoFactX dataset (train split) for geography-based factual reasoning,and translated s1k prompts~\citep{hwang2025learngloballyspeaklocally} for mathematical reasoning. The prompts are machine-translated from English and cover five and ten languages, respectively. We prompt the teachers to generate chains-of-thought and answers for training a student model (details in \cref{app:reasoning}).

\subsection{Evaluation Benchmarks} 
We focus on challenging, multilingual benchmarks that test our models' \textit{generative} abilities and cover tasks of three domains (full details in \cref{app:evals}):

\textbf{Open-ended challenging prompts (Arena)} are sourced from \textit{mArenaHard V.2} \citep{khairi2025lifegivessamplesbenefits} (11 languages).
Quality of generations is measured in terms of win rates as determined by an LLM judge (\texttt{gpt-4o-2024-05-13}) (1) in direct comparison to the commercial \geminiflash{} and \geminipro{} models and (2) in head-to-head comparisons of \fusion{} vs \bon{}.

\textbf{Machine Translation (WMT)} prompts are sourced from \textit{WMT24++}~\citep{deutsch2025wmt24expandinglanguagecoverage,kocmi-etal-2024-findings} (English to 10 languages).
Quality of generations is measured with \textsc{XComet-XL}~\citep{guerreiro-etal-2024-xcomet},
a state-of-the-art multilingual translation evaluation metric.

\textbf{Reasoning} evaluations target the reasoning fine-tuning mix and include  the GeoFactX test split~\citep{hwang2025learngloballyspeaklocally} (5 languages) and math problems from \textit{MGSM} (11 languages incl. English)~\citep{shi2022language}. Both are evaluated in terms of accuracy of the final answers, and we additionally inspect reasoning quality for GeoFactX, following~\citep{hwang2025learngloballyspeaklocally}.

\section{Results}\label{sec:results}
\subsection{Test-time Scaling}\label{sec:results_test-time}

\begin{figure}[t]
   \centering
   \includegraphics[width=\textwidth]{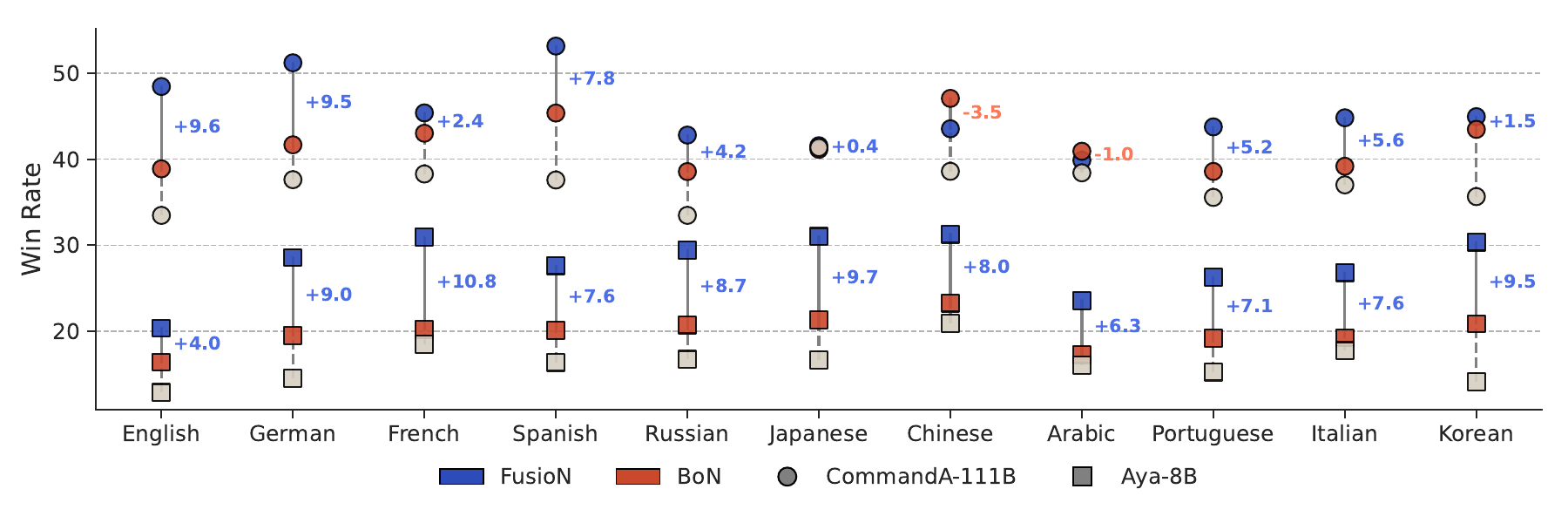}
   \caption{\textbf{Test-time scaling with $N=5$}: \fusion{} raises win rates against \geminipro{} on Arena across languages. It largely outperforms \bon{} with the same set of samples, for both \aya{} and \command{} models. Gray markers indicate greedy baseline performance.}
\label{fig:fusion_arena_aginast_gemini_tts}
\end{figure}

\textbf{\fusion{} brings substantial improvements in multilingual open-ended generation tasks.}
We evaluate both \command{} and \aya{} on Arena when scaling test-time compute (\cref{sec:tts_setup}) and comparing gains from using \fusion{} vs \bon{}. The results in \cref{fig:fusion_arena_aginast_gemini_tts} show  significant gains in win-rate against \geminipro{} across both languages and models. For \aya{} we see impressive jumps in win-rate of  up to +10.8\% in French. Similarly, \fusion{} outperforms \bon{} for \command{} in 9 out of 11 languages. Surprisingly, in cases like German (+9.5\%) and Spanish (+7.8\%) the gains from using \fusion{} on only 5 samples allow \command{} to \textit{win over \geminipro{}} (absolute win-rate $>$ 50\%), the top model in Arena. 
This special case, where fusor and sampling model are identical, \fusion{} can be seen as a form of very effective self-refinement~\citep{ranaldi-freitas-2024-self}.
The gains from \fusion{} are also consistent at different scales (number of samples), tasks and in direct comparison which we  investigate deeper in \cref{sec:analysis} and \cref{sec:extended_results}.

\begin{figure*}[h]
    \centering
    \includegraphics[width=\textwidth]{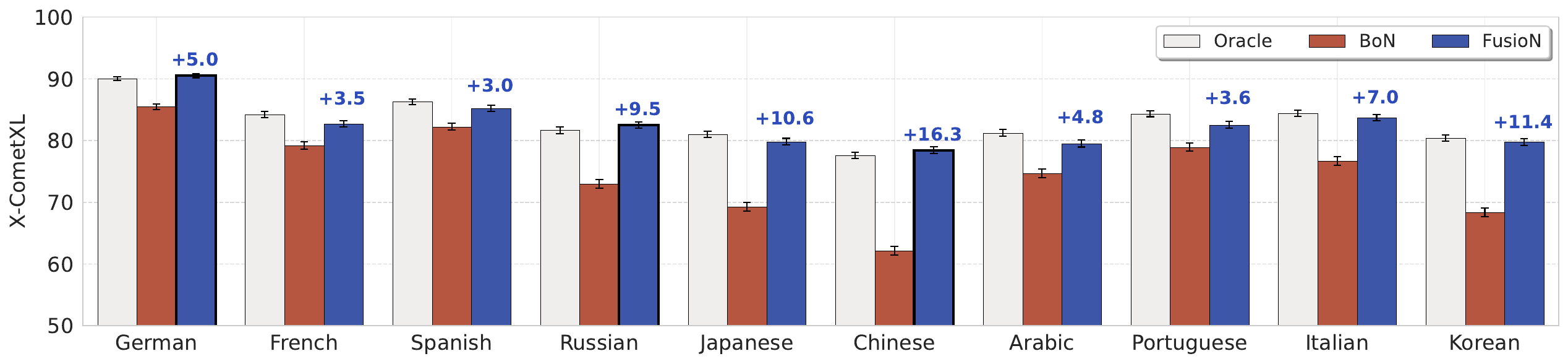}
    \caption{\fusion{} vs \bon{} vs \textsc{Oracle} (the highest scoring sample according to the ground truth) in Translation, error bar show std-err. Bars with bold border (German, Russian and Chinese) are cases where \fusion{} is outperforming the \textsc{Oracle}}
    \label{fig:across_tasks_fusion}
\end{figure*}

\textbf{Synthesis beats selection in machine translation.} 
When testing on WMT we can use the reference translation to score each candidate generation against it with the task metric \comet{}. We can thus identify the ``oracle'' among our samples, and compare its quality to the quality of samples selected by \bon{} with its (imperfect) RM, or the sample synthesized by \fusion{}. 
\Cref{fig:across_tasks_fusion} shows the comparison for $N=5$ generations from \command{} sampled at temperature $\tau=0.7$ for the WMT24++ test set. 
\fusion{} outperforms \bon{} with large margins across languages, reaching differences of +11.4 in Korean. 
More importantly, \textit{\fusion{} outperforms the \textsc{Oracle}} selection in the German, Russian and Chinese translation with gains of +0.8 in the latter, a meaningful improvement in terms of \comet{} scores. This confirms the utility of our proposed synthesis framework of aggregation. Instead of treating generations as competitors in a zero-sum game, we should treat them as collaborators whose strengths can be integrated.

\subsection{Synthetic Data Generation}\label{sec:results_downstream}

\newcommand{\lightgreen}{\cellcolor{green!0}}
\newcommand{\mediumgreen}{\cellcolor{green!0}}
\newcommand{\darkgreen}{\cellcolor{green!0}}

\begin{table}[t]
\centering
\resizebox{0.9\textwidth}{!}{%
\begin{tabular}{lcccccccccccc|c}
\toprule
& & \large ar & \large de & \large en & \large es & \large fr & \large it & \large ja & \large ko & \large \large pt & \large ru & \large zh & \textit{Avg} \\
\midrule
\multirow{3}{*}{\rotatebox{90}{\Large Arena}} & \textbf{\bon{}} & 43.9 & 43.1 & 42.7 & 43.3 & 44.5 & 44.2 & 43.6 & 45.1 & 43.4 & 43.7 & 44.8 & \textit{43.8} \\
&  \textbf{\fusion{}} & \textbf{45.1} &\textbf{ 44.3} & \textbf{48.0} & \textbf{46.2} & \textbf{48.3} & \textbf{48.4} & \textbf{43.8} & \textbf{48.4} & \textbf{45.0} & \textbf{45.2} & \textbf{46.3} & \textit{\textbf{46.3}} \\
&  $\Delta$
& \lightgreen +1.2 
& \lightgreen +1.2 
& \darkgreen +5.3 
& \mediumgreen +2.9 
& \mediumgreen +3.8 
& \darkgreen +4.2 
& \lightgreen +0.2 
& \mediumgreen +3.3 
& \lightgreen +1.6 
& \lightgreen +1.5 
& \lightgreen +1.5 
& \mediumgreen \textit{+2.5} \\
\midrule
\multirow{3}{*}{\rotatebox{90}{ \Large WMT}}  & \textbf{\bon{}} & 73.8 & 90.9 & - & 86.4 & 83.5 & 85.6 & 81.6 & 81.7 & 85.1 & 83.0 & 78.6 & \textit{83.0} \\
& \textbf{\fusion{}} & \textbf{74.6} & \textbf{91.2} & - & \textbf{87.2} & \textbf{84.3} & \textbf{86.2} & \textbf{83.1 }& \textbf{82.8 }& \textbf{85.5} & \textbf{83.5} &\textbf{ 79.8} & \textit{\textbf{83.8}} \\
& $\Delta$ 
& \mediumgreen +0.8* 
& \lightgreen +0.3* 
& - 
& \mediumgreen +0.8* 
& \mediumgreen +0.8* 
& \mediumgreen +0.6 
& \darkgreen +1.5* 
& \darkgreen +1.1* 
& \lightgreen +0.4 
& \lightgreen +0.5 
& \darkgreen +1.2* 
& \mediumgreen \textit{+0.8} \\
\bottomrule
\end{tabular}%
}
  \caption{\textbf{Downstream evaluation} of \bon{}/\fusion{}-fine-tuned 111B models on Arena (\% win rate against \geminiflash{}) and WMT (\comet{}, en$\rightarrow \cdot$): \fusion{} outperforms \bon{} consistently across tasks and languages. 
  * indicates significance for WMT results according to \texttt{comet-compare} (paired t-test and bootstrap resampling~\citep{koehn-2004-statistical}). 
    The baseline starts with an average score of 22.8\% for Arena, and 82.0\% for WMT.
    }
    \label{tab:downstream_ufb}
\end{table}

\textbf{\fusion{} yields consistent multilingual gains with downstream impact.} We compare generation and translation quality of the model fine-tuned on \fusion{}-generated data with the model trained on \bon{}-generated data in \cref{tab:downstream_ufb} (see \cref{sec:extended_results} for 7B results). All hyperparameters, prompts and teacher outputs are identical for both variants. Given that we only change the way we aggregate the samples, we find surprisingly notable and consistent improvements of \fusion{} over \bon{}, across languages and the two tasks. On average, the model fine-tuned on fused generations yields  \comet{} scores +0.8 higher on WMT24++, a delta that can be expected to represent human preferences with around 73.6\% accuracy, according to estimates in~\citep{kocmi2024thresholds}.\footnote{\url{https://kocmitom.github.io/MT-Thresholds/}}  Similarly, \fusion{} improves win-rates against \geminiflash{} by +2.5\% over \bon{}. With only minimal intervention in the data generation phase, the results reveal a remarkable downstream impact, underscoring the powerful ripple effect that even modest improvements in data generation can achieve.The 7B model finetuned with \fusion{} outperforms the one finetuned with \bon{} on WMT, but not Arena as we discuss in \cref{sec:extended_results}.

\begin{figure}[h]
    \centering
    \includegraphics[width=0.5\linewidth]{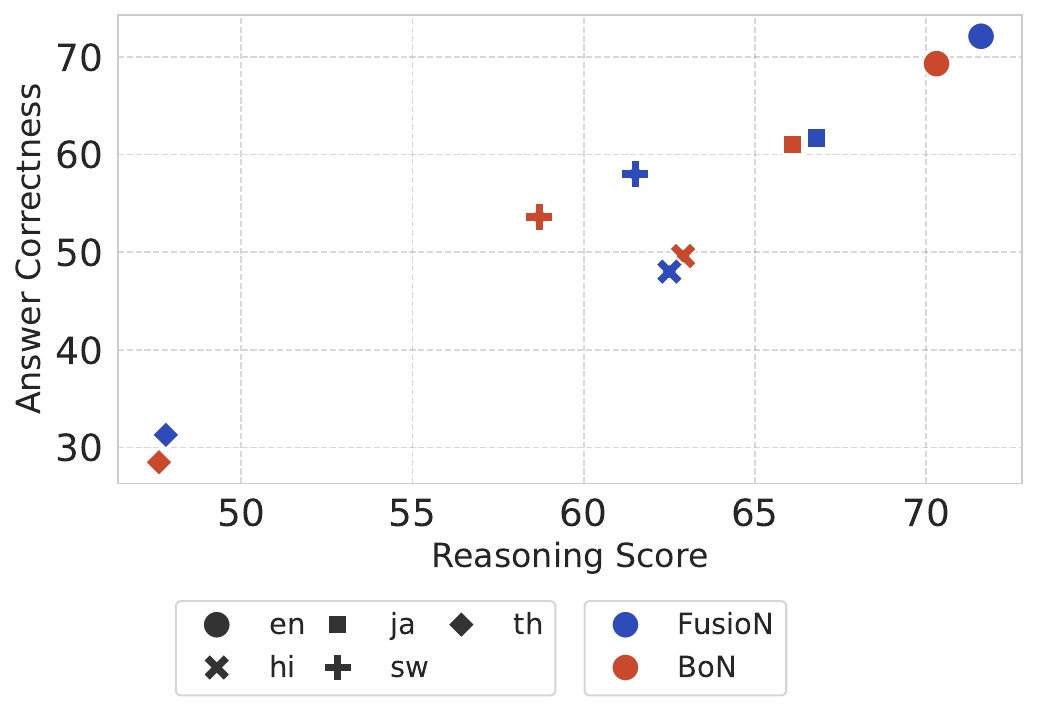}
    \caption{\textbf{Downstream evaluation on multilingual factual reasoning} on the {GeoFactX} test set. \fusion{} outperforms \bon{} notably in both reasoning quality and answer correctness in 4/5 languages.}
    \label{fig:geofactx_scatter}
\end{figure}

\textbf{\fusion{} leads to better multilingual factual reasoning} \Cref{fig:geofactx_scatter} demonstrates how the model fine-tuned on \fusion{} outputs outperforms the model fine-tuned on \bon{} in terms of answer correctness and reasoning score across four out of five languages, with a minor regression in Hindi. The gap for lower-resourced languages, Swahili and Thai, is particularly large (e.g. \fusion{} scores +2.8\% higher than \bon{} in answer correctness and reasoning score for Swahili), showing that we can even more effectively exploit teacher diversity there. 
The fine-tuned models do not only outperform base model (by +5.2\% in answer correctness on average for \bon{}, +7.1\% for \fusion{}), but also the fusor model (by +1.7\% and +3.5\%, respectively, see full results in \cref{tab:geofactx}). This validates our hypothesis that we can effectively leverage the wisdom of the crowd without being bounded by the model that performs the fusion (see also \cref{sec:extended_results}).
It is worth noting that this holds even for the languages that the fusor model (\command{}) officially does not support (Swahili and Thai). 
On MGSM, however, we found some cases where \fusion{} scores below \bon{}, which we discuss in~\cref{app:synth_extra}.

\section{Analysis}\label{sec:analysis}

Our results reveal consistent improvements across setups and languages with an out-of-the box fusor and a small set of samples. To find out, \emph{where and how} \fusion{} is working, we conduct a range of ablations, diving deeper into specific sub-questions.

\begin{figure}[h]
    \centering
    \includegraphics[width=0.5\linewidth]{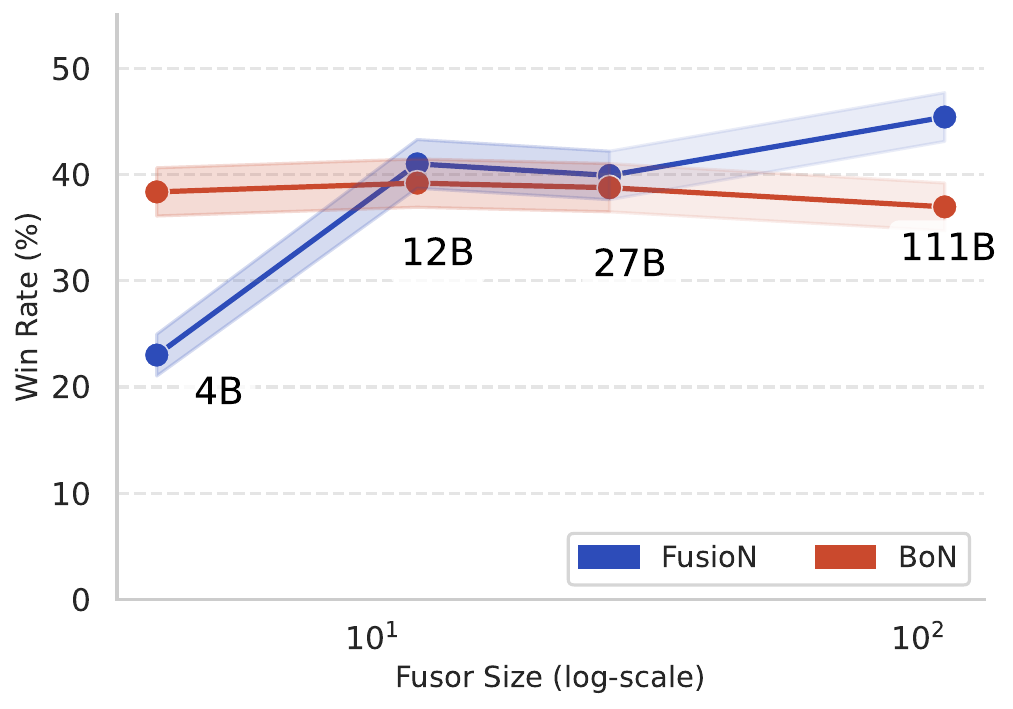}
    \caption{\textbf{Size of the fusor matters}: Small LLMs might serve well as scalar judges in \bon{}, but generative fusion capabilities get unlocked at larger scale, here measured in win-rates on Arena, averaged across languages, shaded areas represent std-err.}
    \label{fig:fusor_ablation}
\end{figure}

\textbf{What makes the fusor work?} 
In \cref{fig:fusor_ablation} we approach this question from two angles: (i) the scale of the fusor (number of parameters), and (ii) how the fusor model is utilized. We evaluate the size effect by varying the fusor from the \textit{4B} Gemma-3 to the \textit{111B} \command{} measuring the resulting average win-rate of test-time scaling on Arena.
We find that for \fusion{} (blue) \textbf{a larger scale  is needed for the fusor to work out of the box}. Importantly, \fusion{} continuously benefits from increasing the scale of the fusor as we see an increase in win-rates of +5.5\% as we go from the \textit{27B} fusor to a \textit{111B} fusor.
When we use the same fusor models as a \textit{rater} in \bon{} (red) (prompt in \cref{app:prompts}), smaller models fare better, but these gains vanish at scale, which aligns with the observation that even the strongest generative models such as \geminipro{} are still outperformed by classifier RMs on classic reward scoring benchmarks~\citep{malik2025rewardbench2advancingreward}.
Overall, \fusion{} utilizes the judge capabilities at larger scale more effectively than \bon{}.

\begin{figure}[h]
    \centering
   \includegraphics[width=0.65\textwidth]{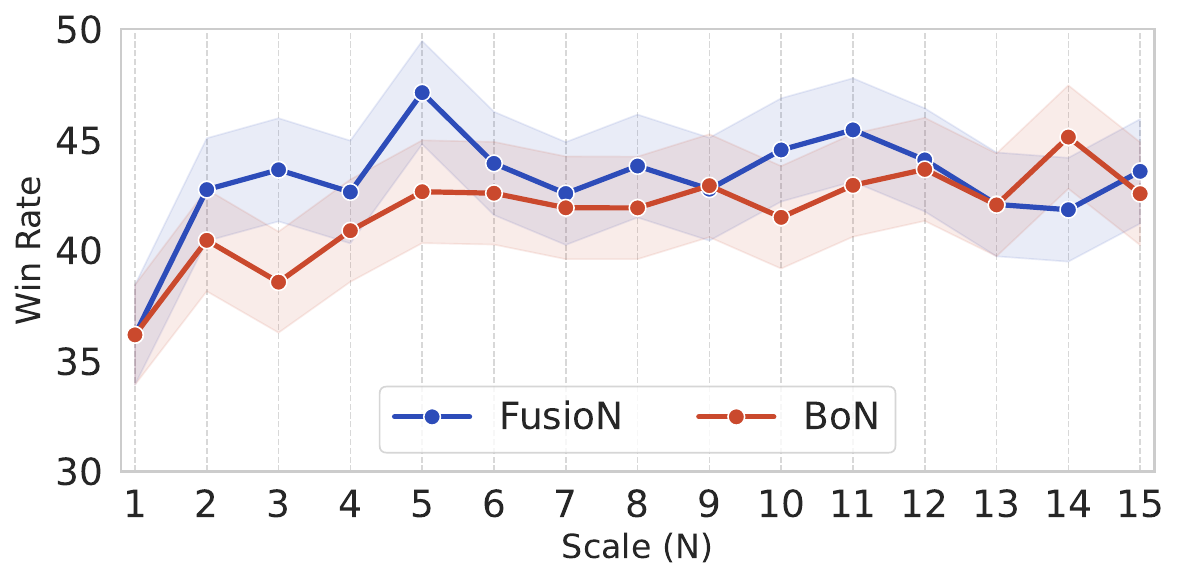}
    
       \caption{\textbf{Scaling test-time budget}: 
       Win-rates are shown against \geminipro{} on 4 languages from m-ArenaHard-v2.0. Shaded areas are average std-err across languages. }
   \label{fig:fusion_arena_scaling_tts}
\end{figure}

\textbf{Which method is more sample-efficient?}
We compare the sample efficiency of \bon{} and \fusion{} under the same test-time scaling budget. 
In \cref{fig:fusion_arena_scaling_tts} we measure win-rates on Arena across four languages (See \cref{sec:extended_results} for language breakdown). We observe that \fusion{} is more efficient at the lower scales ($N<10$), improving  win-rate against \geminipro{} by +6\% with only $N=2$, where \bon{} needs twice more samples to achieve similar gains. 
Gains for both methods plateau beyond $N=7$, but \fusion{} consistently makes fuller use of each generated sample, making \textbf{\fusion{} the more efficient choice for low-budget scaling}.
Note that \bon{} requires $N$ independent samples, which are parallelizable, while \fusion{} encodes all samples together. 
Despite this, \fusion{} shines at small $N$, making every token count and turning even a few samples into high-quality, integrated solutions. 
With an efficient long-context implementation, it can achieve strong scaling performance while fully leveraging the diversity in the sample pool.

\begin{figure}
\begin{floatrow}

\capbtabbox{%
  \begin{tabular}{@{\hspace{0.5em}}l@{\hspace{0.5em}}c@{\hspace{1em}}l@{\hspace{0.5em}}c@{\hspace{0.5em}}}
\toprule
\textbf{\#} & \textbf{Candidate Pool} & \textbf{Method} &  \textbf{WR}  \\
\midrule
0 & CMD: 1 Sample                                & - & 57.9  \\
\midrule
1 &  \multirow[c]{3}{*}{\makecell{All 5 Teachers}} & \fusion{} & 65.4  \\
2 &                                                & \bon{}    & 61.0 \\
\cline{3-4}
\addlinespace[1pt]
3 &                                                & Fusor=DS & 63.9  \\
\midrule
\addlinespace[0.8pt]
4 & \multirow[c]{2}{*}{{Weaker Pool}}              & \fusion{} & 65.0  \\
5 &                                                & \bon{}    & 60.9 \\
\midrule
6 & \multirow[c]{2}{*}{{Smaller Pool}}             & \fusion{} & 62.9  \\
7 &                                                & \bon{} & 60.2 \\
\midrule

8 &  \multirow[c]{2}{*}{\makecell{DS: 5 Samples}} & \fusion{} & 59.0  \\
9 &                                               & \bon{}    & 58.9  \\
\bottomrule
\end{tabular}
}{%
  \caption{\textbf{Ablation on pool size and diversity}: win-rates (\textit{WR}) vs. \geminiflash{} 
  using 1k random UFB samples, averaged over 10 languages.
\textit{DS}: \deepseek{}. \textit{CMD}: CommandA.}\label{tab:ufb_ablations}
}
\ffigbox{%
  \includegraphics[width=0.48\textwidth]{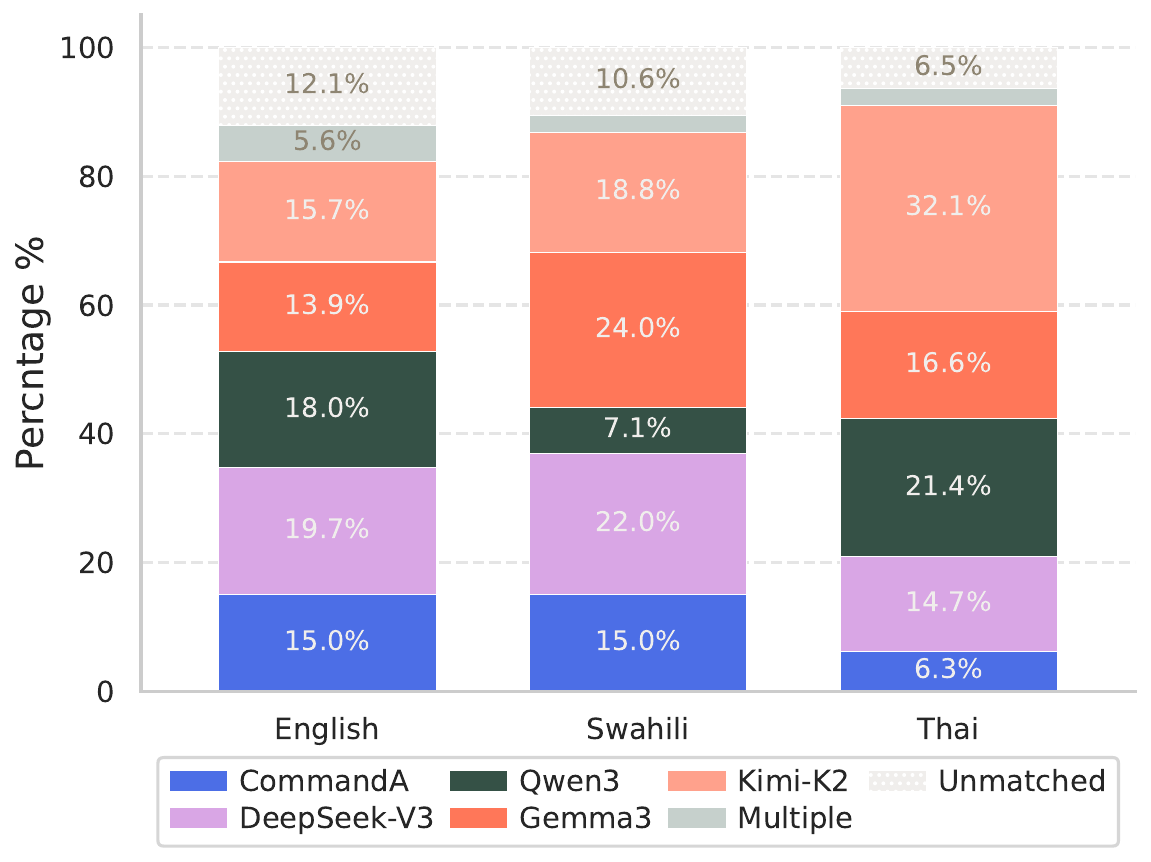}
}{%
  \caption{\textbf{Diverse teacher contributions}:   Analysis of teacher contributions to the final output of \fusion{} on a subset of GeoFactX (50 samples per language) across English and languages not supported by the fusor.}
\label{fig:fusion_contribution_ood_lang}
}

\end{floatrow}
\end{figure}

\textbf{How is synthetic data quality affected by the fusor and teacher pool?} The quality of the synthetic data generated is dependent on the quality of the pool of the samples and the fusor used.  We measure quality of the data averaged across 10 languages using win-rates against \geminiflash{} for 1k examples of UFB, and report in \cref{tab:ufb_ablations} how modifications to the teacher pool and fusor affect data quality. 
Across all modifications we see that \fusion{}-generated synthetic data is of higher quality compared to the \bon{} data with +4.4\% in the default setup with all five teachers. 
Even when we perform \fusion{} with---on this benchmark---weaker \deepseek{} (based on reward scores from our internal RM) (\#3), we see only a small drop in quality while still out-performing \bon{}. When we replace \gemmabig{} with \gemmasmall{} in the teacher pool (weaker teacher pool) (\#4+\#5), both methods are minimally affected---however, using a smaller pool of only four teachers (without \kimi{}, \#6+\#7) affects \fusion{} proportionally more, but it still wins over \bon{}. 
If we sample only from a single teacher (here \deepseek{}, \#8+\#9) win-rates drop substantially, highlighting the importance of diversity in the teacher pool.
Overall, these ablations show that \fusion{} is \textbf{more robust under weaker ensembles} than \bon{}.

\begin{figure}[h]
   \centering
   \includegraphics[width=1\textwidth]{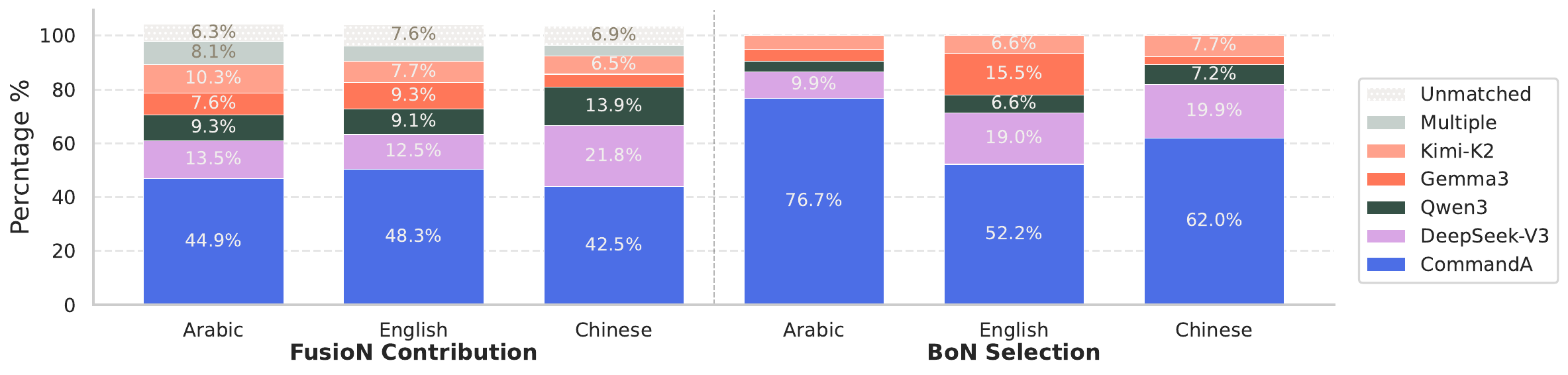}
\caption{\textbf{Contributions to the final generation:} Analysis of different teachers in the pool contributing to the final output of \fusion{} or \bon{}. We look at outputs on subset of UFB (50 samples per language) with a 5-teachers pool. }

\label{fig:fusion_contribution_vs_bon}
\end{figure}

\textbf{How does \fusion{} balance its pool?} We track the sources of contribution in \fusion{} by surface-level sequence matching (details and more bias probes in \cref{app:contribution}, with an example in \cref{fig:example_fused_generation}). 
We see that a large proportion of the \fusion{} output is directly taken from the teachers' outputs, forming a coherent synthesis. 
\textbf{Only a small fraction of words is \textit{unmatched}}, where the fusor adds ``glue'' or reformulates teacher outputs. 
In \cref{fig:fusion_contribution_vs_bon}, we compare contribution to fused outputs and selections of \bon{} on a random sub-sample of UFB. 
While both methods show similar high-level preferences (favoring \command{} the most and \gemmabig{} the least), \fusion{} integrates even the less preferred ones. 
Finally, we inspect the contributions in the GeoFactX data, because it contains languages not officially covered by the fusor (\command{}). \Cref{fig:fusion_contribution_ood_lang} shows that \fusion{} remains robust, with its preferences shifting to utilize \gemmabig{}{} the most.

\begin{table}[t]
\centering
\begin{tabular}{@{}cccc@{}}
\toprule
Model & \makecell{Reward\\Score} &  \makecell{Answer\\Correctness} &  \makecell{Language\\Correctness} \\
\midrule
 \fusion{}     & \textbf{7.86} & \textbf{58.56} & 98.86   \\
 \bon{}        & 7.11          & 49.77          & \textbf{99.05}   \\
\addlinespace[1pt]
\midrule
 CommandA     & 6.30          & 40.52          & \textbf{99.61} \\
 DeepSeek     & 6.43          & \textbf{48.89} & 98.99          \\
 Qwen-3       & 6.46          & 41.06          & 99.58          \\
 Gemma-3      & 6.52          & 41.89          & 92.11          \\
 Kimi-K2      & \textbf{6.64} & 46.62          & 99.56          \\
\bottomrule
\end{tabular}
\caption{\textbf{Data analysis} for teachers and aggregation outputs on GeoFactX samples, averaged across languages.}
\label{tab:synth_data_analysis}
\end{table}

\textbf{Is the fusor putting a ceiling on the quality?} In \cref{tab:synth_data_analysis}, we compare the outputs of \fusion{}, \bon{}, and all teachers on various metrics on the GeoFactX train set. For \textit{answer correctness}, \fusion{} achieves the highest accuracy with 58.6\%, despite the fusor (\command{}) scoring the lowest. \fusion{} also obtains the highest \textit{reward score}, from our internal RM, on the GeoFactX mix---the very metric \bon{} is optimizing. We note that \fusion{} has a low 98.9\% \textit{language correctness}, likely due to our fusion prompt being English-only (\cref{app:prompts}), but leave studying these effects to future work. These results show that \textbf{ \fusion{} is not limited by the fusor} and is in fact more dependent on the sample pool.

\textbf{Opportunities for strengthening \fusion{}.} 
During test-time scaling we find that \fusion{} benefits English more than other languages. It is likely that the skills that are required to perform successful \fusion{} are not evenly present in all languages. Although we do find gains of \fusion{} over \bon{} in unsupported languages of the fusor for GeoFactX, \bon{} might be the safer choice for cross-lingual transfer to lower-resource languages~\citep{hong2025cross}, while generative model capabilities are still lacking behind.
We also found more mixed results when testing on MGSM (\cref{app:synth_extra}), which might indicate that close-ended tasks are either just not well suited to be addressed by generative ensembling, or that the fusor would need specialized training for such a specialized domain that RMs are usually well trained on.

\section{Related Work}\label{sec:related}

\textbf{Learning from Ensembles}
The principle of learning from ensembles has led to advances in many areas of machine learning, and can be integrated into training LLMs in various forms: 
For example, \citet{huang2024ensemble} fuse multiple models via their output probabilities, \citet{lee-etal-2023-ensemble} learn from a consensus of multiple teachers in self-instructing~\citep{wang-etal-2023-self-instruct}, and \citet{wan2024knowledge} propose a continual pretraining objective for knowledge distillation from multiple teachers.
In this work, we focus on \textit{integrative output ensembling}, where we simply provide a LLM (the fusor) the ensemble of outputs as input to integrate their strengths into a \emph{fused} output.

\textbf{Synthesis-based Ensembling}
Our approach can be seen as an instance of \textit{Mixture-of-Agents} (MoA) ~\citep{wang2024mixtureofagents}, a framework where multiple agents organized in layers iteratively enhance the output. Our approach stands out through simplicity: We show that \fusion{} becomes effective already in a \textit{single aggregation step} with a single fusor, even in diverse and challenging setups, thereby constituting an attractive alternative for \bon{}, which is---thanks to its simplicity---a much more widely adopted framework than MoA. \\
\textit{LLM-Blender}~\citep{jiang-etal-2023-llm} follows a similar idea, but requires two separate modules, one for pairwise ranking, and one for fusing top-ranking outputs. In contrast to our work, this framework operates on the basis of pairwise comparisons (which require training a specialized model), while we argue that the fusor should receive \textit{all outputs at once} to best comparatively evaluate them.\\
Other contemporaneous related works also require training such specialized aggregator modules
~\citep{qi2025learningreasonparallelsamples, zhao2025majorityrightrltraining, li2025draftsanswersunlockingllm}, while our approach is effective \textit{without any training}. These works focus primarily on verifiable tasks like math and code targeting RL  or reasoning. For such specialized scenarios with expert models available, 
\citet{li2025rethinkingmixtureofagentsmixingdifferent} warn that MoA might not be sufficiently robust to lower-quality inputs. 
For the more diverse generative evaluation scenarios that we are targeting, however, we find that \fusion{} is fairly robust with respect to the teacher pool (\cref{sec:analysis}), and sampling from a single teacher---the proposed solution by \citet{li2025rethinkingmixtureofagentsmixingdifferent}---performs significantly worse. 
\citet{jayalath2025computeteacherturninginference} find that fused single-teacher roll-outs can nevertheless provide valuable supervision in RL training, even without any fusor training. 
Overall, our work fits nicely in a stream of very recent developments discovering new possibilities of synthesis as part of the inference process. 
Even though the idea of \fusion{} is so intuitive and shared among recent works, our work advances the understanding of the inner workings and limitations of this principle. We show that implemented even in its simplest form, it brings gains in highly diverse applications for both at test-time and for driving model supervision.

\textbf{Test-time Scaling} Our approach can also be cast as  combination of parallel and sequential test-time-scaling~\citep{welleck2024from, Snell2025ScalingLT}, with $N$ parallel steps and one refinement step. 
\citet{inoue2025widerdeeperscalingllm}
formulate this combination as a search problem where in a each step either more samples can be requested, or existing ones can be revisited. This poses an interesting avenue for future work, where \fusion{} operates with adaptive compute (rather than a fixed $N$+1) based on each input sample. This flexibility might be needed for attempts to mimic human cognitive processes more closely~\citep{zhang2024buildingspecializedgeneralistai}.

\textbf{Synthetic Data Generation}
In the development of multilingual LLMs in particular, synthetic data generation has played a core role to reduce language disparities.
For example, two recent models Apertus~\citep{hernándezcano2025apertusdemocratizingopencompliant} and EuroLLM~\citep{martins2025eurollm9btechnicalreport}, rely on EuroBlocks,\footnote{\url{https://huggingface.co/datasets/utter-project/EuroBlocks-SFT-Synthetic-1124}} a collection of synthetic fine-tuning data obtained from various sources and individual teachers. 
Such synthetic data has also been key in improving mathematical reasoning, both monolingually~\citep{muennighoff2025s1simpletesttimescaling} and multilingually \citep{lai-nissim-2024-mcot,hwang2025learngloballyspeaklocally}.
Involving and ensembling multiple generations from either the same or multiple teachers in the process, as we study here, is still underexplored. 
For Llama 3, ~\citet{grattafiori2024llama3herdmodels} report using rejection sampling (i.e. \bon{}) for multilingual data generation. 
For Aya Expanse, ~\citet{dang2024ayaexpansecombiningresearch} report routing samples to  multiple teachers~\citep{lu-etal-2024-routing} via multilingual \bon{} as proposed in ~\citep{odumakinde-etal-2025-multilingual}, a strategy also adopted for building Tower+~\citep{rei2025towerbridginggeneralitytranslation}.

Overall, our work complements very recent advancements discovering collaborative synthesis at inference, enhancing understanding of its benefits and limitations. Even in its simplest form, our approach demonstrates gains across diverse applications, including test-time scaling and model supervision.

\section{Conclusion}
Our work thoroughly investigates and challenges the to-date standard practice of \bon{} in test-time scaling and synthetic data generation. Our experiments strongly support replacing it by \fusion{} in these scenarios to make most of the costs that are already incurred from generating and evaluating multiple samples.
Across a range of challenging multilingual tasks, \fusion{} consistently outperforms traditional winner-takes-all approaches like \bon{}, delivering higher-quality outputs, greater sample efficiency, and stronger downstream performance. 
Importantly, \fusion{} leverages the strengths of multiple models, even when some are weaker, showing robustness and adaptability.
These results highlight a shift in how we should think about evaluating and utilizing LLM generations: rather than measuring quality monolithically, embracing their polylithic nature allows us to integrate diverse strengths, unlock latent potential, and achieve improvements that were previously inaccessible through selection alone. 
\fusion{} points toward a more effective and sustainable paradigm for leveraging the collective capabilities of today's leading LLMs.

\section*{Acknowledgments}
We thank our colleagues for their help in various stages of this project:
Wei-Yin Ko, Kylie He and David Mora for the help with post-training, Kelly Marchisio for her advice regarding benchmarking, Thomas Euyang for the beautiful illustration, Madeline Smith for the help with communications, Sara Hooker and Ye Shen for feedback in discussions in early stages of the project, and the remaining Cohere Labs for their helpful feedback throughout all iterations of this project. Furthermore, we would like to thank Jaedong Hwang for sharing  data and evaluation code for the synthetic reasoning experiments.

\section*{Ethics Statement}
Training on synthetic data comes with inherent risks of propagating and amplifying biases~\citep{ahn-etal-2022-knowledge,shimabucoro-etal-2024-llm,mohammadshahi2025whats}. We hope that by increasing diversity in the teacher pool, we can reduce model-specific biases to propagate (as opposed to learning from one teacher only), and prevent loss of diversity the generated data~\citep{briesch2024largelanguagemodelssuffer}.
Regardless, we cannot strictly protect the student model from adversarial teachers, probably even less so with \fusion{} than \bon{} because they might be more prone to prompt injections. Our tests revealed robustness with respect to the quality of the teacher pool (\cref{sec:analysis}), but we have not tested truly adversarial inputs. We rely on the user to verify teacher suitability and potentially add any sanity checks. In contrast to \bon{}, the \fusion{} framework allows for flexible instructions that could include e.g., a constitution~\citep{bai2022constitutionalaiharmlessnessai} or specific safety guidelines. In practice, \fusion{} could also be prepended with a hard filter for unsafe or lowest-quality samples (e.g. language compliance via language identification), so that the undesired information does not even get to the aggregation stage. \\
We also perform additional analyses for typical LLM judge biases in \cref{app:contribution}, and find no evidence for self-preference, but a slight position bias, i.e. the fusor preferring samples that it is presented first more than those that come later. \\
We would also like to emphasize that any use of such ensembling needs to respect all terms of use and licenses of the individual teachers, which lies in the responsibility of the user.

\section*{Reproducibility Statement} 
Fusor and teacher models that we use in this work are publicly available (\cref{sec:experiments}), as well as the prompts for fine-tuning. We transparently report prompts and instruction templates for LLM evaluation (\cref{app:prompts}), and benchmark metric implementations (\cref{sec:experiments}.
Where models are not public (student model in the experiments on synthetic data generation, and reward model), we report scores on public benchmarks that allow to anchor our experiments. 
The data generation pipeline that we describe in detail in \cref{app:reasoning} is not perfectly reproducible due to inherent randomness in the sampling process. Therefore, we release synthetic data for \bon{} and \fusion{} where licenses allow\footnote{
UFB: \url{https://huggingface.co/datasets/CohereLabs/fusion-synth-data-ufb} \\
GeoFactX: \url{https://huggingface.co/datasets/CohereLabs/fusion-synth-data-geofactx}\\
S1KX: \url{https://huggingface.co/datasets/CohereLabs/fusion-synth-data-s1kx}
}.
In addition, we follow the recommended practice for generative multilingual LLM evaluations~\citep{kreutzer2025dj} and release our pairwise evaluations that rely on LLM judges\footnote{
Test-time: \url{https://huggingface.co/datasets/CohereLabs/fusion-pairwise-evals-test-time-scaling}\\
Finetuning: \url{https://huggingface.co/datasets/CohereLabs/fusion-pairwise-evals-finetuned}
}.
\bibliography{custom}

\newpage
\appendix
\section{Prompts}\label{app:prompts}


\begin{table}[]
    \centering
    \begin{tabular}{p{12cm}}
    \toprule
         Based on the provided Instruction and Generated Texts in \texttt{language}, fuse them into a better generation that combines the strength of each of them. Do so in two steps:

First, compare the Generated Text with a focus on what sets them apart in terms of content, language quality and responsibility, highlighting strengths and weaknesses. 
Second, fuse them into a new final generation that combines the best aspects of each of them while avoiding the weaknesses.

The fused generation should be adequately responding to the instruction, sound natural to a native speaker, and be focused on conveying the most relevant and accurate information in a responsible and ethical way.

\underline{Output Format}

Comparison: (short explanation of the strengths and weaknesses of each generation)

Answer: [[ The final fused generation ]]

\underline{Context}

\textbf{Instruction}

\texttt{prompt}

\textbf{Generated Texts}

\texttt{generations}

Please analyse the Generated Texts, discarding any unsafe or unethical generations and provide your fused text. Remember to stick to the requested Output Format, providing first a short explanation and then putting the final fused generation inside double brackets [[]].\\
    \bottomrule
    \end{tabular}
    \caption{Prompt used for \fusion, including \texttt{placeholders}. Generations are randomly shuffled and enumerated, presented one per line. }
    \label{tab:fusion_prompt}
\end{table}

\begin{table}[]
    \centering
    \begin{tabular}{p{12cm}}
    \toprule

Please act as a fair judge. Based on the provided Instruction and Generated Text, analyse the Generated Text and provide a 1-5 integer score. The given instruction is in \texttt{language} and the response should also be in \texttt{language}
Your selection should be based on your judgment as well as the following guidelines for each possible score:

1. The Generated Text is unintelligibly written (incomplete sentences, leaps in logic, flagrant mechanical errors) or has majorly incorrect or unverifiable information.

2. The Generated Text is occasionally difficult to understand, dotted with minor factual or mechanical errors, or missing crucial formatting elements.

3. The Generated Text expresses useful information, is readable, has no factual errors, and has no more than a minor mechanical error or two. Though it may be informative to those unfamiliar with the subject matter, it is not overly insightful, engaging, or likely to hold up to expert scrutiny.

4. The Generated Text clearly expresses useful information at an expert level, is readable, and has no factual or mechanical errors. It could just use a quick adjustment with tone or length.

5. The Generated Text clearly expresses useful information at an expert level, is readable, has no factual or mechanical errors, and is the perfect length and tone with regard to the prompt.

\underline{Output Format}

Analysis: xxx
Answer: [[ SCORE ]] (this should be an integer from 1-5 and nothing else; the score should be enclosed in double brackets as indicated)

\underline{Evaluation Information}

\textbf{Instruction}

\texttt{message}

\textbf{Generated Text}

\texttt{generation}

Please analyse the Generated Text and provide a 1-5 integer score according to the guidelines. Remember to stick to the requested Output Format, providing analysis and putting your final score (an INTEGER in 1-5) inside double brackets [[]].\\

    \bottomrule
    \end{tabular}
    \caption{Prompt used for BoN with generative models, including \texttt{placeholders}}
    \label{tab:gen_bon_prompt}
\end{table}

\begin{table}[]
    \centering
    \begin{tabular}{lp{9cm}}
    \toprule
    Task & Prompt \\
    \midrule
       MGSM (en) & Solve this math problem. Give the reasoning steps before giving the final answer on the last line by itself in the format of "Answer:". Do not add anything other than the integer answer after "Answer": \\
       WMT24++  &  You are a professional \texttt{src\_lang} to \texttt{tgt\_lang} translator, tasked with providing translations suitable for use in \texttt{tgt\_lang} (\texttt{tgt\_country}). Your goal is to accurately convey the meaning and nuances of the original \texttt{src\_lang} text while adhering to \texttt{tgt\_lang} grammar, vocabulary, and cultural sensitivities. Produce only the \texttt{tgt\_lang} translation, without any additional explanations or commentary. Please translate the following \texttt{src\_lang} text into \texttt{tgt\_lang} (\texttt{tgt\_country}):

        \texttt{source\_text}\\
         \bottomrule
    \end{tabular}
    \caption{Instruction prompts used for evaluation, including task-specific \texttt{placeholders}. MGSM prompts are taken from the \href{https://github.com/openai/simple-evals/blob/main/mgsm_eval.py}{simple-evals} library, we only list the English one here but use them in the respective target languages.}
    \label{tab:prompts}
\end{table}

We provide the prompt used by the fusor in \cref{tab:fusion_prompt} . We use the same prompt across all tasks, setups, fusors and languages. \Cref{tab:gen_bon_prompt} shows the pormpt for using the fusor model as scaler rater. We also provide the \textit{English Version} of the instruction prompts used in our evaluation in \cref{tab:prompts}.

\section{Fine-tuning Hyperparameters}\label{app:hyperparams}
 We train the 111B baseline on the synthetic data generated from our UFB mix with a batch size of 16, cosine decay with peak learning rate of 5e-6 using Adam optimizer across 64 Nvidia H100 GPUs for 250 steps. For the extended mix (UFB and Math+GeoFactX) we use the same hyperparameter with increased number of steps of 323. We train the 7B models on the UFB mix with 16 GPUs with the same parameters.

\section{Synthetic Reasoning Data}\label{app:reasoning}

\citet{hwang2025learngloballyspeaklocally} build two datasets for improving multilingual reasoning abilities: \textit{s1k-X} for multilingual mathematical reasoning and \textit{GeoFactX} for geography-based multilingual factual reasoning.
The multilinguality stems from automatic translation of the prompts, of the s1k dataset from ~\citep{muennighoff2025s1simpletesttimescaling}, and of synthetically created English prompts that designed to cover a variety of regions.
For s1k-X the reasoning traces and answers from \texttt{Qwen-2.5-Instruct 72B} in s1k are also translated
(via the Google Translate API). This has some undesired side effects where the mathematical notation or the answer formatting gets corrupted, e.g. with white spaces around \LaTeX{} math symbols.
For both datasets, we only work with the translated prompts (and ignore the provided single-teacher outputs), and use our pool of teachers to generate multilingual responses.
For analysis and for evaluation, we use the human-verified answers provided in the GeoFactX dataset as ground truth for the evaluation of accurateness of answers.
For s1k-X, a correctness analysis of the data is hindered by the inconsistencies in format for the (translated) silver answers of Qwen, which makes the extraction of answers non-trivial. We leave this analysis for future work.

For evaluation of fine-tuned models on the test split on GeoFactX, we follow the procedure in ~\citep{hwang2025learngloballyspeaklocally}. We prompt a LLM judge (here \gpt{}, deviating from ~\citep{hwang2025learngloballyspeaklocally} which uses Qwen, as we wanted avoid self-bias) to score the reasoning traces for quality.  We compare the final answer in the generation against the list of the correct answers provided in the task following the implementation by~\citet{hwang2025learngloballyspeaklocally}, and also verify the language of the response according to their  implementation.\footnote{\url{https://github.com/jd730/M2A/tree/main}} 

\section{Evaluation}\label{app:evals}
We describe our set of evaluation benchmarks in more detail.

\textbf{mArenaHard V.2} (short: \textit{Arena}):\footnote{\url{https://huggingface.co/datasets/CohereLabs/m-ArenaHard-v2.0}} This data contains 498 translated challenging prompts from ArenaHard v2.0\footnote{\url{https://github.com/lmarena/arena-hard-auto/tree/main/data/arena-hard-v2.0}} across 23 languages~\citep{khairi2025lifegivessamplesbenefits}. Quality of generations is measured in terms of win rates in direct comparison to the commercial \geminiflash and \geminipro models in addition to head-to-head comparison of \fusion{} vs \bon{}. We mainly compare against \geminipro{} in the test-time scaling environment where we use production ready LLMs with extra compute. In the synthetic data generation environment, we benchmark weaker LLMs fine-tuned on a small synthetic dataset, hence we switch to \geminiflash{}. The pairwise comparison is as done by a LLM judge, here \gpt. We focus on a subset of 11 languages: English (en), German (de), French (fr), Spanish (es), Russian (ru), Japanese (ja), Chinese (zh), Arabic (ar), Korean (ko), Portuguese (pt), and Italian (it).

\textbf{WMT24++} (short: \textit{WMT}):\footnote{\url{https://huggingface.co/datasets/google/wmt24pp}} This dataset contains translation problems sourced from the WMT 2024 machine translation shared task~\citep{kocmi-etal-2024-findings} expanded to more languages~\citep{deutsch2025wmt24expandinglanguagecoverage}. Quality of generations is measured with \textsc{XComet-XL},\footnote{\url{https://huggingface.co/Unbabel/XCOMET-XL}} a state-of-the-art multilingual translation evaluation metric~\citep{guerreiro-etal-2024-xcomet}. We use the prompt in \cref{app:prompts} and we focus on translating from English to the following languages: Arabic (ar), German (de), Spanish (es), French (fr), Italian (it), Japanese (ja), Korean (ko), Portuguese (pt), Russian (ru), Chinese (zh).

\textbf{MGSM}: This benchmark contains 250 mathematical problems at grade-school level in 11 languages (bn, de, en, es, fr, ja, ru, sw, te, th, zh), originally translated from English ~\citep{shi2022language}. We prompt models to think step by step before outputting the final answer, following the \texttt{simple-evals} implementation,\footnote{\url{https://github.com/openai/simple-evals/tree/main}}. The evaluation metric is the accuracy of the final answer. 

\textbf{GeoFactX} We follow the prompting and evaluation process recommended by ~\citet{hwang2025learngloballyspeaklocally} and evaluate reasoning traces and final answers with an LLM judge and against gold answers, respectively (details in \cref{app:reasoning}).

\section{Additional Results}

\subsection{Test-time Scaling}\label{app:extra_testtime}
\begin{figure}[h]
   \centering
   \includegraphics[width=\textwidth]{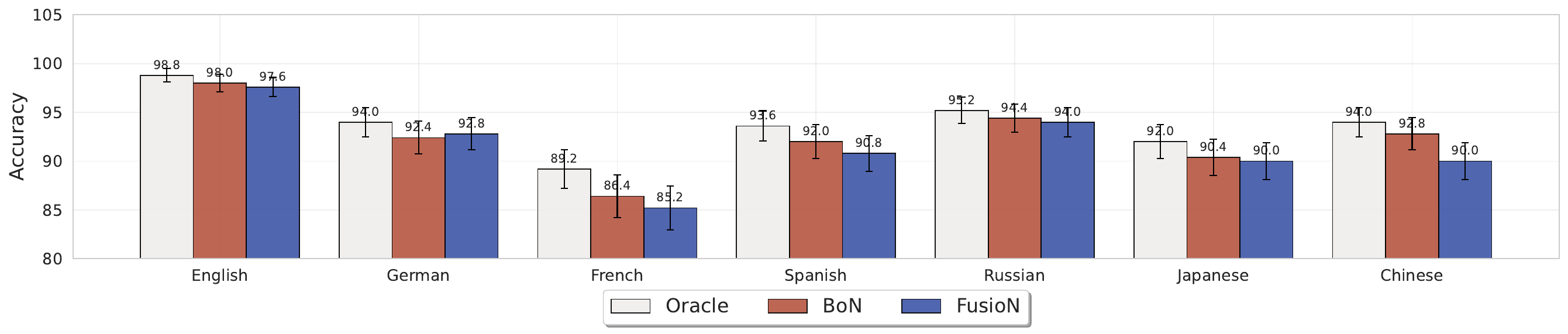}
\caption{\textbf{Test-time Performance on MGSM}.  We find that \bon{} has the best performance across languages, showing that \fusion{} might be the ideal fit for math task compared \bon{} with a specialized RM. Both method performance is close to the \textit{Oracle}. Error bars show std-err.} 
\label{fig:tts_on MGSM}
\end{figure}

Continuing our exploration of \fusion{} in test-time scaling, we look at performance across language on MGSM math benchmark (across a subset of 7 languages) in \cref{fig:tts_on MGSM}. We find that \fusion{} performs well in this as it is relatively close to \textit{Oracle} performance. However, \bon{} has a slight but consistent edge over \fusion{} in all languages (except German) in this task.

\subsection{Synthetic Data Generation}\label{app:synth_extra}
\begin{figure}[h]
   \centering
   \includegraphics[width=0.8\textwidth]{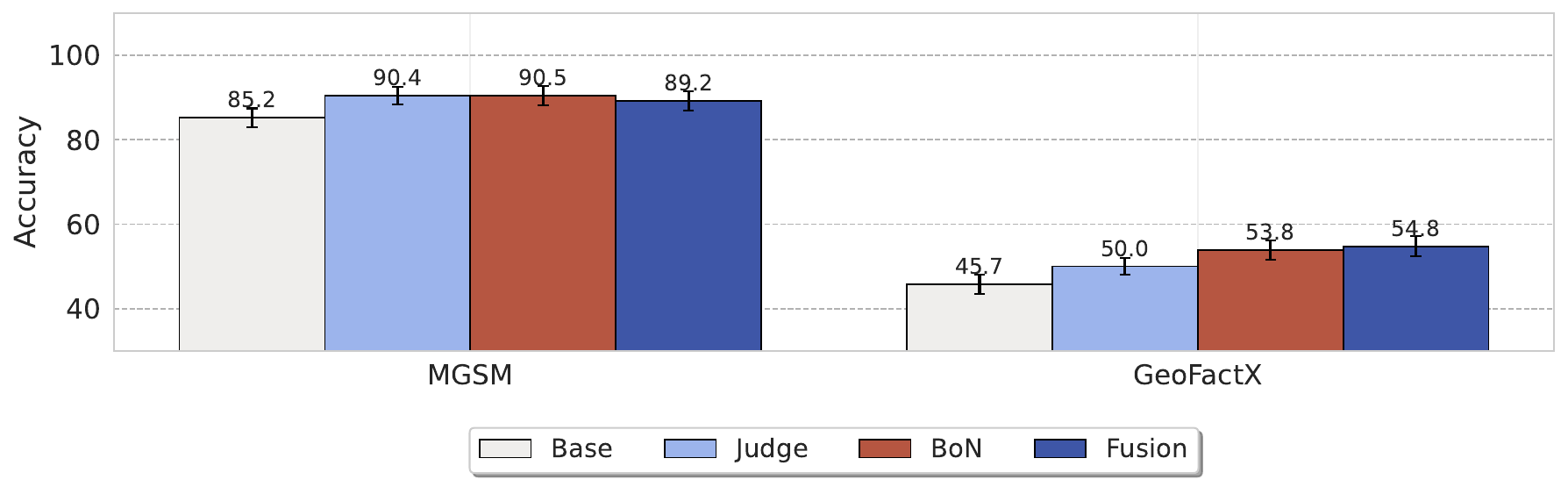}
\caption{\textbf{Comparison of downstream performance on close-ended tasks MGSM (11 languages) and GeoFactX (5 languages)}. We find while on math \bon{} has the best performance, \fusion{} has higher accuracy on GeoFactX, outperforming the fusion judge as well. Error bars show std-err averaged across languages.} 
\label{fig:sft_on_close_ended_task}
\end{figure}

\textbf{Reasoning tasks} \Cref{fig:sft_on_close_ended_task} compares the performance on the two reasoning tasks for the models trained on \fusion{} vs \bon{} in relation to the performance of the \textsc{Judge} model, i.e., the fusor model (\command{}) and also the \textsc{Base} model that we start fine-tuning from.
For MGSM, we find that the \fusion{} accuracy lags behind both \bon{} -1.2\% and the \textsc{Judge} performance of -1.5\%, indicating that while \fusion{} is overall beneficial for improving downstream math accuracy (+3.6\% above \textsc{Base}), it is not the optimal choice in this case (similar to the test-time scaling experiment in \cref{app:extra_testtime}. Other works \cite{qi2025learning} have shown that it is possible to train specialized LLMs to perform synthesis in the math reasoning domain, which might help to remedy that. But the facts that (1) \bon{} does not improve the performance of the fine-tuned model beyond the fusor's performance, and that (2) the baseline model performs already surprisingly strong, make us wonder whether these results could also be due to the interplay between data seen in prior training steps, in fine-tuning, and also in the fusor model. Since the s1k dataset is quite popular, it might have been part of training (in English) of the fusor and the baseline already. 
For the factual QA domain we see in stark contrast, that with clearly unseen data, \fusion{} effects stand out more. The model fine-tuned on \fusion{} achieves the best accuracy with +1.0\% gains over \bon{} and an impressive +4.8\% compared to fusor \textsc{Judge}.

\section{Contribution Analysis}\label{app:contribution}
\textbf{Measuring contributions} We inspect \fusion{} outputs and compare them with the teacher outputs with string matching. While this does not capture semantic rephrasings, it does give us an idea how much the fusor can directly copy and paste blocks of the teacher outputs. Parts of the \fusion{} output that we cannot directly find in any teacher's output, we mark as ``unmatched''---this is where we might have some close semantic matches  or also just some ``glue'' work to connect parts from different teachers.
The matching procedure works as follows:
\begin{enumerate}
    \item Finds all matching blocks between the fused string and each teacher string. We make use of the \texttt{difflib} library\footnote{\url{https://docs.python.org/3/library/difflib.html}} and use their \texttt{SequenceMatcher} to detect the longest contiguous matching subsequences.
    \item Resolve attribution for each character: Retrieve matching blocks that cover it, assume the teacher with the longest match wins. If there is a tie: mark it as ``multiple''. If there is no match, mark it as ``unmatched''.
    \item Calculate contribution statistics for each teacher: Count how many characters of the fused generation it was attributed to.
\end{enumerate}

\begin{figure}[h]
   \centering
   \includegraphics[width=0.8\textwidth]{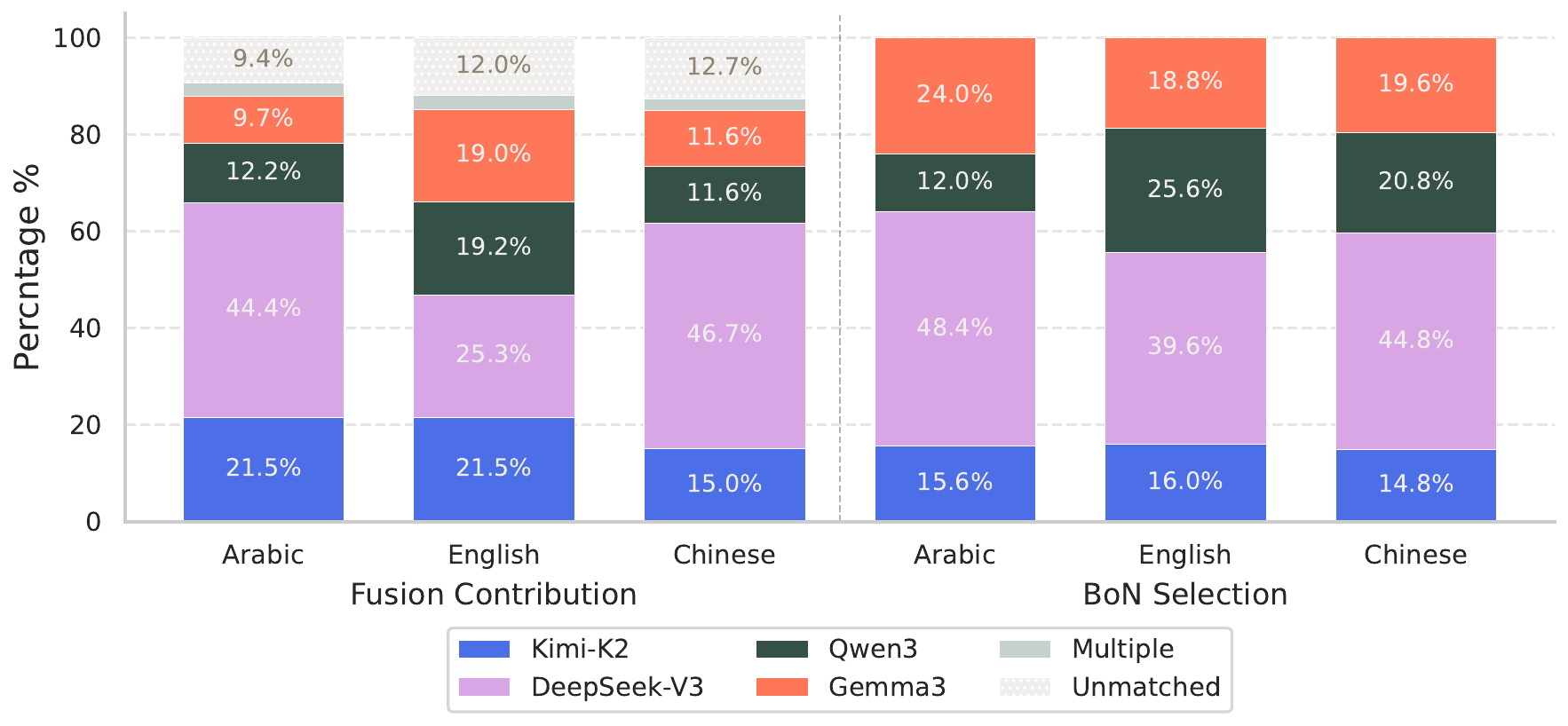}
\caption{\textbf{\fusion{} contributions without fusor in the teacher pool:} Analysis of different teachers in the pool contribution to the final output of \fusion{} when the fusor model is \textit{not} in the pool. We look at outputs on subset of UFB (50 samples per language)}
\label{fig:fusion_contribution_without_command_in_pool}
\end{figure}

\textbf{Disentangling fusor and teacher pool} The resulting contribution statistics can be compared with \bon{} selections that chooses one teacher for each sequence. In \cref{fig:fusion_contribution_without_command_in_pool} we do this on a small subset of the UFB data mix covering languages Arabic, English and Chinese. We look at pool of teachers that does not include the fusor (\command{}) to study the effect of the fusor self-bias. Similar to what we found in \cref{fig:fusion_contribution_vs_bon} where \fusion{} and \bon{} had their highest contribution from \command{} (the teacher that we now removed), in \cref{fig:fusion_contribution_without_command_in_pool} the methods also have same preference, agreeing on \deepseek{} as their favorite (previously second-ranking when \command{} was in the pool). 
This consistent preference lets us conclude that \fusion{} does not suffer from self-bias with the fusor able to reliably find the best samples in the pool, whether or not the fusor samples is one of them.

\begin{figure}[htbp]
    \centering
    \begin{subfigure}[b]{0.45\textwidth}
        \centering
        \includegraphics[width=\textwidth]{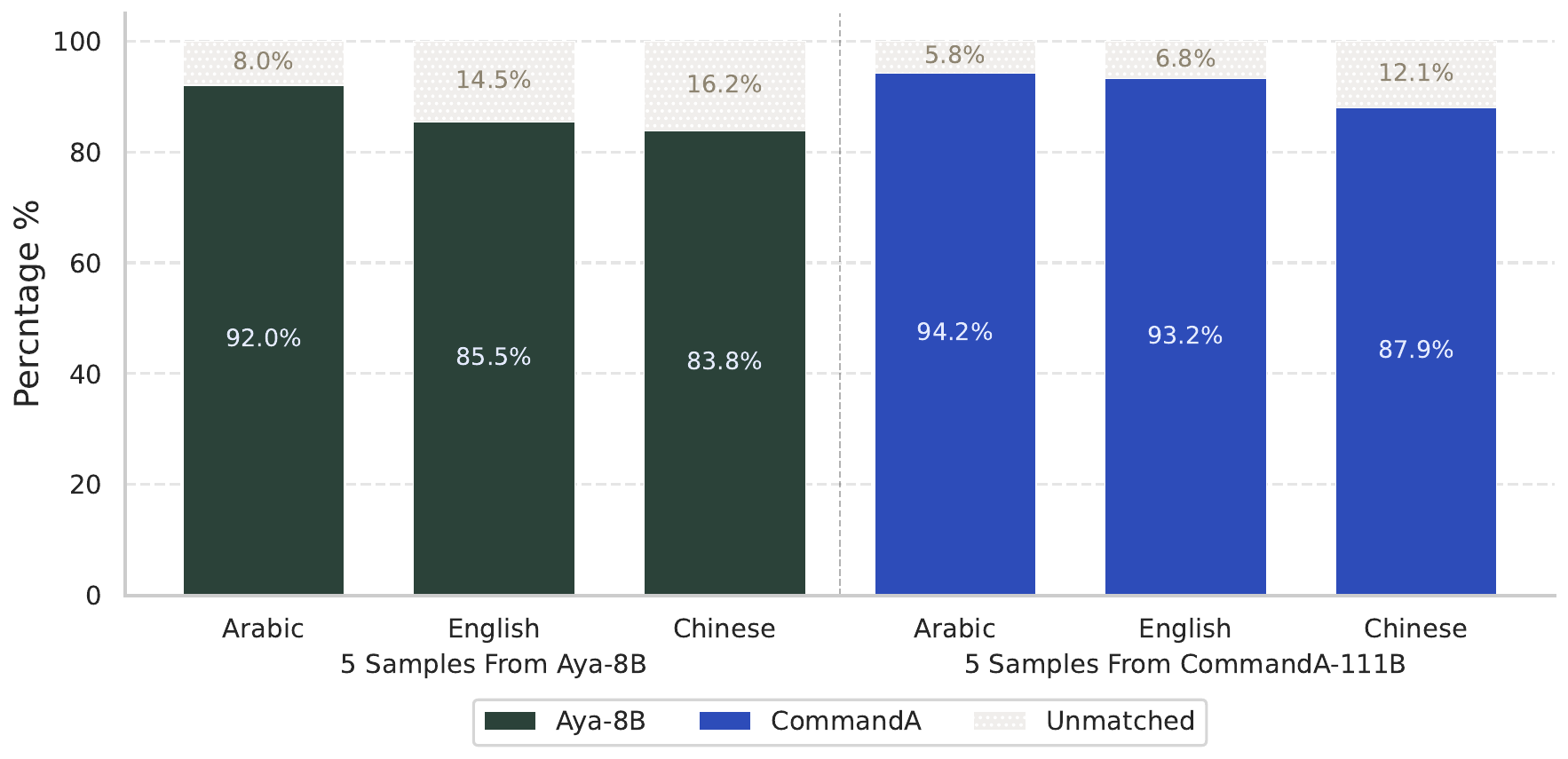}
        \caption{Across \aya{} and \command{}}
        \label{fig:fusion_contribution_tts_only_grouped}
    \end{subfigure}
    \hfill
    \begin{subfigure}[b]{0.45\textwidth}
        \centering
        \includegraphics[width=\textwidth]{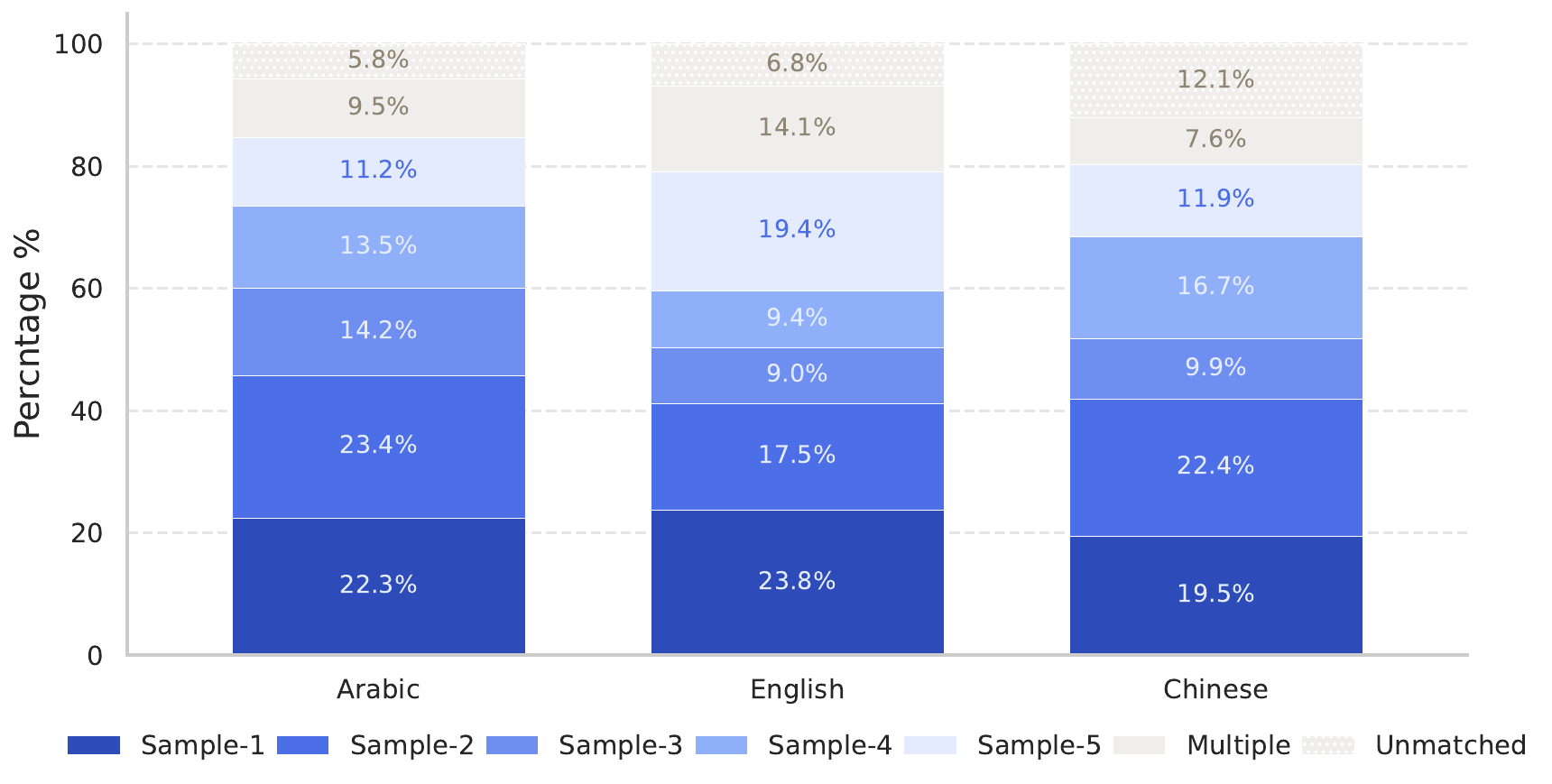}
        \caption{Across samples from \command{}}
        \label{fig:fusion_contribution_tts_only_per_sample}
    \end{subfigure}
    
    \caption{\textbf{\fusion{} contributions in test-time scaling:} Analysis of how different samples in test-time scaling contribute to the final output of \fusion{}: (a) with different candidate models (b) based on samples order. We look at \fusion{} outputs on subset of mAreanHard-v2. (50 samples per language)}
    \label{fig:fusion_contribution_test_time_scaling}
\end{figure}

\textbf{Contributions in test-time scaling} In \cref{fig:fusion_contribution_test_time_scaling} we perform the contribution analysis on the test-time scaling setup, where the samples in the pool are coming from a single model. First, we examine the effect of changing this single model on the \fusion{} preference. We are mostly interested in the case when the fusor is much larger than the candidate model as one would assume the fusor may opt to replace all of the weaker model outputs with its own preference in the fusion which would result in higher \textit{unmatched} rate. However, in \cref{fig:fusion_contribution_tts_only_grouped} we see that for both the small candidate model \aya{} and the larger one \command{} the fusor has a small  \textit{unmatched} rate.  Albeit larger for \aya{}, it demonstrate the fusor outputs are almost always more than 80\% from the content in the samples. 

\textbf{Position bias} We consider another type of possible bias in \cref{fig:fusion_contribution_tts_only_per_sample}, where we visualize the contribution analysis based on the order of the samples in the \fusion{} prompt. The samples are always shuffled before being formatted  (see \cref{tab:fusion_prompt}) and we analyze the order based on what the order the fusor sees. We find while not large in magnitude, across the three languages, the fusor outputs have the highest contribution ratios in the first and second samples.

\definecolor{bgcolor}{HTML}{FAF9F6}
\definecolor{bordercolor}{HTML}{E2DFD2}
\definecolor{colortitle}{HTML}{020508}

\begin{figure}[]
\centering
\resizebox{0.80\linewidth}{!}{%
  \begin{tcolorbox}[colback=bgcolor, 
  colframe=bordercolor,
  rounded corners,
  title=Can a comedian's use of audience interaction enhance their delivery and stage presence?,
  coltitle=colortitle,
  boxrule=0.6pt,
  left=2mm, right=2mm, top=1mm, bottom=1mm
  ]
    \input{fusion_examples/all_comedy}
  \end{tcolorbox}
}
\caption{An excerpt of the outputs of teachers to the \fusion{} example show in \cref{fig:example_fused_generation}}
\label{fig:example_fused_generation_teachers}
\end{figure}

\begin{figure}[ht]
\centering
\resizebox{\linewidth}{!}{%
  \begin{tcolorbox}[colback=bgcolor, 
  colframe=bordercolor,
  rounded corners,
  title=Can a comedian's use of audience interaction enhance their delivery and stage presence?,
  coltitle=colortitle,
  boxrule=0.6pt,
  left=2mm, right=2mm, top=1mm, bottom=1mm
  ]
    \input{fusion_examples/comedy_long}
  \end{tcolorbox}
}
\caption{Contribution analysis for a sample \fusion{} output with text colored based on the respective source: 
\textcolor[HTML]{2D4CB9}{CommandA},
\textcolor[HTML]{9B60AA}{DeepSeek-V3}, 
\textcolor[HTML]{2B4239}{Qwen3}, 
\textcolor[HTML]{CA492D}{Gemma3}, 
\textcolor[HTML]{8E8572}{multiple}, and 
\textcolor[HTML]{BBB1A4}{unmatched}. Individual teacher generations are in \cref{fig:example_fused_generation_teachers}.
}
\label{fig:example_fused_generation}
\end{figure}

Finally in \cref{fig:example_fused_generation_teachers} and we provide an example prompt with excerpts from teachers outputs, and in \cref{fig:example_fused_generation} we show the full \fusion{} output color coded according to our contribution analysis.

\section{Evaluation Results Comparison} \label{sec:extended_results}

\begin{table}
\centering
    \begin{tabular}{lccc|ccc}
    \toprule 
     & \multicolumn{3}{c|}{\textbf{mArenaHard v2.0} } & \multicolumn{3}{c}{\textbf{WMT24++}} \\
   \textbf{Language}  & \multicolumn{3}{c|}{ (win-rate, in \%)} & \multicolumn{3}{c}{(\comet{}; en$\rightarrow \cdot$)} \\
        (+region for WMT)  & \textbf{\bon{} }& \textbf{\fusion{}} & $\Delta$ & \textbf{\bon{} }& \textbf{\fusion{}} & $\Delta$\\
            \midrule
            ar (SA) & \textbf{18.3} & 15.6 & -2.7 & 65.4 & \textbf{66.3} & +0.9 \\
            de (DE) &      16.8     & 16.8 &  0.0 & 86.8 & \textbf{87.1} & +0.3 \\ 
            en      & \textbf{14.9} & 14.1 & -0.8 &  -   &     -         &   -  \\
            es (MX) & \textbf{19.6} & 17.6 & -2.0 & 81.0 & \textbf{81.5} & +0.5 \\ 
            fr (FR) & \textbf{22.4} & 17.2 & -5.2 & 77.4 & \textbf{77.5} & +0.1 \\ 
            it (IT) & \textbf{19.4} & 17.2 & -2.2 & 80.1 & \textbf{80.4} & +0.3 \\
            ja (JP) & 17.0 & \textbf{19.5} & +2.5 & 70.8 & \textbf{72.1} & +1.3 \\
            ko (KR) & \textbf{17.9} & 14.0 & -3.9 & 72.0 & \textbf{72.5} & +0.5 \\ 
            pt (PT) & \textbf{16.9} & 16.6 & -0.3 & 79.8 & \textbf{80.2} & +0.4 \\
            ru (RU) & 12.8 & \textbf{19.0} & +6.2 & \textbf{74.8} & 74.5 & -0.3 \\
            zh (CN) & 19.8 & \textbf{20.9} & +1.1 & 70.9 & \textbf{71.6} & +0.7 \\
            \midrule
            \textit{Avg} & \textbf{\textit{17.8}} &	\textit{17.1} & \textit{-0.7} & \textit{75.6} & \textbf{\textit{76.4}}	 & \textit{+0.5}\\
            \bottomrule

    \end{tabular}
    \caption{\textbf{Downstream evaluation results on 7B Models} of \bon{}/\fusion{}-fine-tuned 7B models on mArenaHard v2.0  (win rate against \geminipro{} as judged by \gpt{}) and WMT24++ (\comet{}).}
    \label{tab:downstream_ufb_7b}
\end{table}

To evaluate the effectiveness of our method and generated data across different scales, we applied our synthetic data generation and SFT pipeline to a smaller baseline. We followed the same setup as used for the 111B model. The 7B baseline is a base  model that have not undergone any post-training stages. As shown in \cref{tab:downstream_ufb_7b}, the downstream results vary. While \fusion{} remains more effective than \bon{} in WMT24, improving performance in every language, we observed no gains significant over \bon{} in m-ArenaHard-v2.0. This suggests that the smaller model requires more parameter tuning to achieve an optimal setup for SFT to be effective in downstream performance, especially the relatively small size of our synthetic dataset.

\begin{table}[t]
    \centering
    \resizebox{0.8\textwidth}{!}{%
    \begin{tabular}{lccccc|ccccc}
    \toprule
         &\multicolumn{5}{c}{\textbf{ Reasoning Score}} & \multicolumn{5}{c}{\textbf{Answer Correctness}} \\
         &\multicolumn{5}{c}{ (LLM score, in \%)} & \multicolumn{5}{c}{(Accuracy, in \%)} \\
        \textbf{} & \textbf{\bon{}} & \textbf{\fusion{}} & $\Delta$ & \textit{Baseline} & \textit{Fusor} & \textbf{\bon{}} & \textbf{\fusion{}} & $\Delta$ & \textit{Baseline} & \textit{Fusor}\\
        \midrule
        en& 70.3	& 71.6  & +1.3 & 66.5 & 69.2 & 69.3 & 72.1 &	+2.8 & 75.2 & 73.2\\
        hi & 62.9   & 62.5	& -0.4 & 54.3 & 61.6 & 49.6 & 48.0 &	-1.6 & 45.1 & 51.6\\
        ja & 66.1   & 66.8	& +0.7 & 56.7 & 61.6 & 61.0 & 61.7 &	+0.7 & 57.5 & 59.3\\
        sw & 58.7	& 61.5	& +2.8 & 31.7 & 44.8 & 53.6 & 58.0 &	+4.4 & 35.3 & 44.0\\
        th & 47.6	& 47.8  & +1.1 & 34.2 & 39.4 & 28.5 & 31.3 &	+2.8 & 22.8 & 25.6\\
         \midrule 
         \textit{Avg} & \textit{61.1}& \textit{62.2}	& \textit{+1.1} & \textit{48.7} & \textit{55.3} & \textit{52.4}&\textit{ 54.2}	&\textit{+1.8} & \textit{47.2} & \textit{50.7}\\
         
         \bottomrule
    \end{tabular}%
    }
    \caption{\textbf{Downstream evaluation on multilingual factual reasoning}, as measured on the GeoFactX test set. \fusion{} outperforms \bon{} both in terms of reasoning quality and answer correctness, with the exception of Hindi. \textit{Baseline} is model we used to finetune and \textit{Fusor} is the model used for fusing the generations.}
    \label{tab:geofactx}
\end{table}

We also report the detailed breakdown of results on downstream gains in the multilingual factual reasoning benchmark from the finetuned 111B model in \cref{tab:geofactx}. The \textit{Fusor} and \textit{Baseline} help anchor the gain and provide context about the magnitude of the finetuning gains in general. We find that fine-tuning consistently improve performance over the baseline---with one exception in Hindi. More importantly \fusion{} also outperform the \textit{Fusor} in most results.

\begin{figure}[h]
   \centering
   \includegraphics[width=0.75\textwidth]{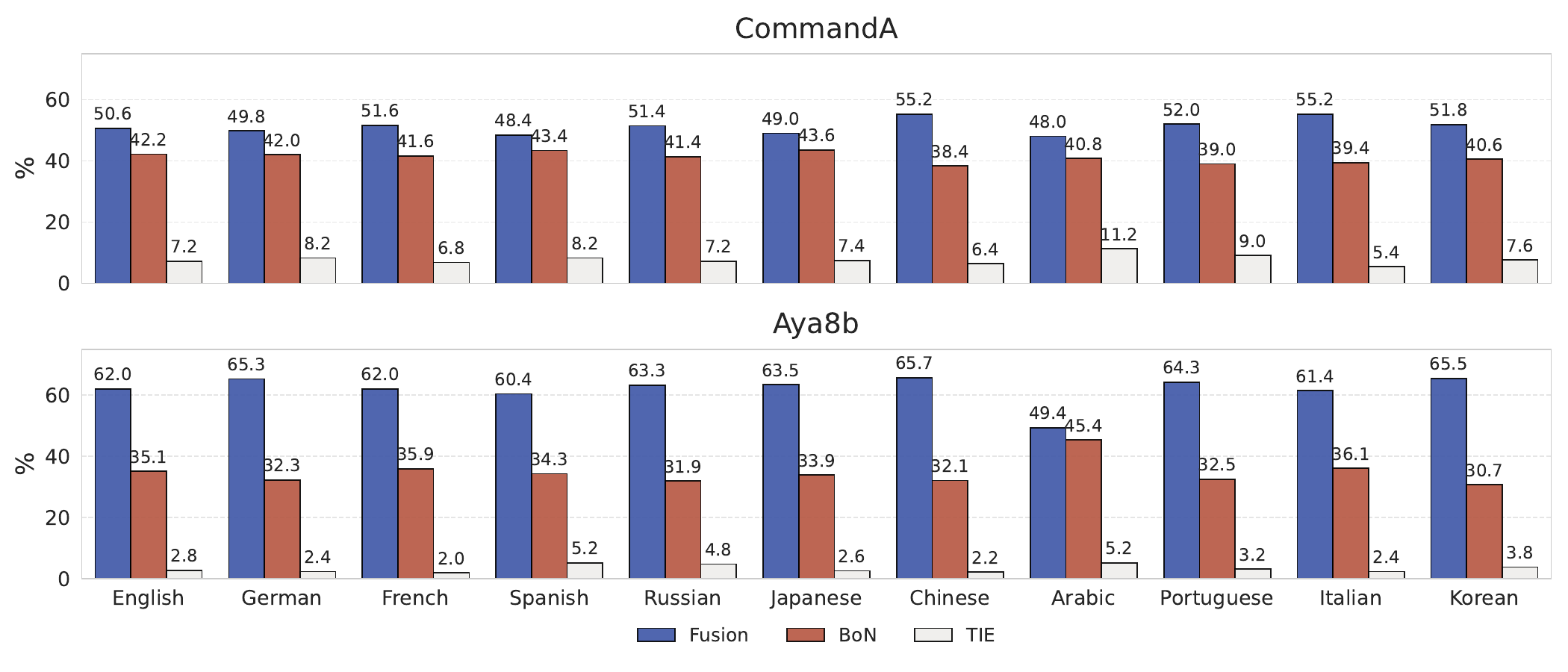}
\caption{\textbf{Test-Time Scaling}: Head-to-head comparison of \fusion{} vs \bon{} on m-ArenaHard-v2, judged by \gpt{}. We consistently see that \fusion{} outperforms \bon{} for both \aya{} and \command{}.}
\label{fig:tts_head_head}
\end{figure}

\begin{figure}[h]
   \centering
   \includegraphics[width=0.75\textwidth]{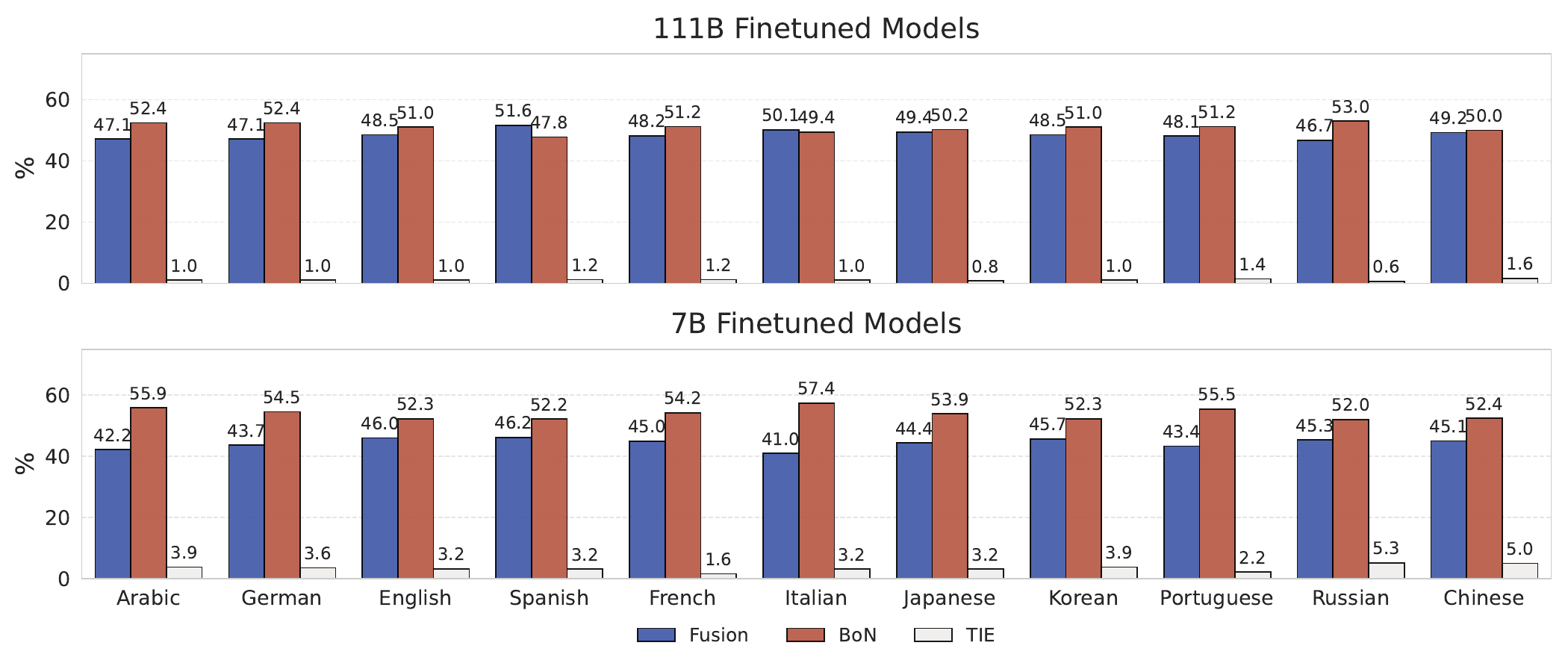}
\caption{\textbf{Synthetic Data Finetuning}: Head-to-head Comparison of the models finetuned with \fusion{} vs \bon{} on m-ArenaHard-v2. We find that variants are mostly on par in the 111B case, while \bon{} is better on the 7B level.}
\label{fig:sft_head_head}
\end{figure}

In \cref{fig:tts_head_head}, we move from comparing against \geminipro{} as our reference and directly evaluate \fusion{} against \bon{} in pairwise head-to-head setup judge by \gpt{}. We use \aya{} and \command{} to generate 5 samples on Arena, and aggregate with either one of our methods. We find significant gain that resemble what we observe in \cref{fig:fusion_arena_aginast_gemini_tts}: \fusion{} outperform \bon{} across languages with large gains in magnitude for \aya{} and lower but also impressive margins in for \command{} (up to +55.2\% win-rate in Italian). 

We perform a similar comparison in \cref{fig:sft_head_head} between the finetuned models at the 111B and 7B. We see varying results across languages in direct head-to-head comparison between the \fusion{} finetuned and \bon{} finetuned 111B models. For the 7B we see that \bon{} scores better across all languages.

\begin{figure}[h]
   \centering
   \includegraphics[width=0.75\textwidth]{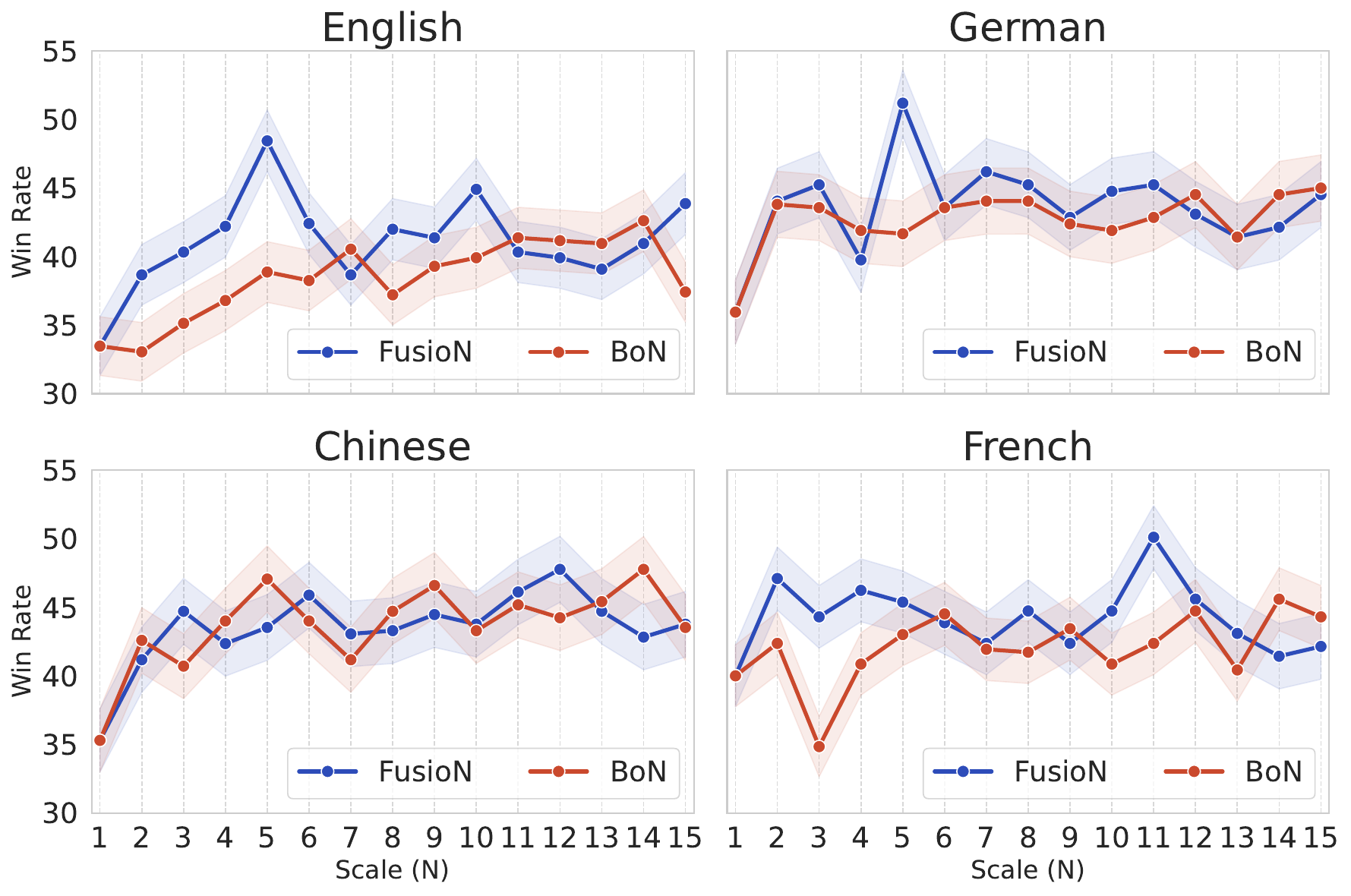}
\caption{\textbf{Sample Efficiency at scale}: test-time scaling win-rates of \fusion{}/\bon{} vs \geminipro{} on m-ArenaHard-v2, judged by \gpt{}. For German and Chinese, both variants perform similarly with larger $N$, while for English and French there is a clear advantage of \fusion{} over \bon{} when working with fewer samples $N<=6$.}
\label{fig:scaling_per_lang}
\end{figure}

In \cref{fig:scaling_per_lang} we provide a breakdown on the scaling plots (win-rates on m-ArenaHard-v2.0 vs \geminipro{}) across 4 languages. We can see that in most languages that \fusion{} grows faster than \bon{}, with magnitudes depending on the language.

\section{LLM Usage Disclosure}
In this paper, we used AI in several auxiliary functions:
\begin{itemize}
    \item Formatting of result tables in \LaTeX{}.
    \item Shortening the text to fit into space limits.
    \item Polishing text by finding English correspondences to our non-English ideas.
    \item Implementation aid for the contribution analysis.
    \item Expansion of our initial list of related works, which we then read and carefully curated into the final related work discussion.
\end{itemize}

\end{document}